\documentclass[journal, 10pt]{IEEEtran}
\usepackage{epsfig,endnotes}
\usepackage[hidelinks]{hyperref}
\usepackage{url}
\usepackage{xcolor,colortbl}
\usepackage{booktabs,caption,fixltx2e}
\usepackage[flushleft]{threeparttable}
\usepackage{wrapfig}
\usepackage{makecell}
\usepackage{float,graphicx}
\usepackage[font=small]{caption}
\usepackage{subcaption}
\usepackage{multicol}
\usepackage{multirow}
\usepackage{float}
\usepackage{tikz}
\usepackage{mwe}
\usepackage{arydshln}
\usepackage{listings}
\usepackage{amsmath,mathtools,amssymb}
\usepackage{array, makecell}
\usepackage{enumitem}
\usepackage{microtype}
\usepackage{verbatim}
\usepackage{setspace}
\usepackage{bookmark}
\usepackage[ruled,linesnumbered]{algorithm2e}

\SetCommentSty{mycommfont}
\usepackage{xcolor,colortbl}
\definecolor{ashgrey}{rgb}{0.7, 0.75, 0.71}
\SetKw{KwBy}{by}
\usepackage{textcomp}
\usepackage[normalem]{ulem}
\newcommand{\ul}{\textit}
\makeatletter
\g@addto@macro{\UrlBreaks}{\UrlOrds}

\newcommand{\encircle}[1]{%
    \tikz[baseline=(char.base)]{\node[shape=circle,draw,inner sep=0pt] (char) {#1};}%
} 

\definecolor{Gray}{gray}{0.85}
\definecolor{Gainsboro}{rgb}{0.86, 0.86, 0.86}
\definecolor{lblue}{rgb}{0.7,1,1}

\definecolor{codegray}{rgb}{0.5,0.5,0.5}
\definecolor{codepurple}{rgb}{0.58,0,0.82}
\definecolor{backcolour}{rgb}{0.95,0.95,0.92}
\usepackage[belowskip=-15pt,aboveskip=0pt]{caption}

\definecolor{codegray}{rgb}{0.5,0.5,0.5}
\definecolor{codepurple}{rgb}{0.58,0,0.82}
\definecolor{backcolour}{rgb}{0.95,0.95,0.92}

 \lstdefinestyle{mystyle}{
    backgroundcolor=\color{backcolour},   
    commentstyle=\color{codegray},
    keywordstyle=\color{magenta},
    numberstyle=\tiny\color{codegray},
    stringstyle=\color{codepurple},
    basicstyle=\ttfamily\scriptsize,
    breakatwhitespace=false,         
    breaklines=true,                 
    captionpos=b,                    
    keepspaces=true,                               
    numbersep=5pt,                  
    showspaces=false,                
    showstringspaces=false,
    showtabs=false,                  
    tabsize=2
}

\makeatletter
\def\endthebibliography{%
  \def\@noitemerr{\@latex@warning{Empty `thebibliography' environment}}%
  \endlist
}
\ifCLASSINFOpdf
\else
\fi

\makeatother

\begin{document}

\title{Leaking Circuit Secrets: Gradient Leakage Attacks on Graph Neural Networks
}

\author{
  \IEEEauthorblockN{Rupesh Raj Karn},  \IEEEauthorblockN{Johann Knechtel} \IEEEauthorblockN{Ozgur Sinanoglu}\\
  \IEEEauthorblockA{Center for Cyber Security, New York University, Abu Dhabi, UAE.}\\
  Email: \{rupesh.k, johann, ozgursin\}@nyu.edu
}

\maketitle

\begin{abstract}
As graph neural networks (GNNs) become standard tools for critical tasks in circuit design and analysis, their security and privacy risks require careful attention.
Here, we present the first comprehensive evaluation of gradient leakage attacks (GLAs) on GNNs in circuit-design and hardware-security tasks, a practical threat that has been largely overlooked.
We assess state-of-the-art (SOTA) GNNs, including GraphSAGE, GCN, GIN, and GAT, trained on standard netlist benchmarks (ISCAS'85, EPFL, and TrustHub), for their fundamental vulnerability to GLAs.
We find that GLAs can expose sensitive information, such as gate types and distinctive properties of hardware Trojans, which may assist adversaries in analyzing logic locking schemes or evading Trojan detection mechanisms.
Our analysis shows that these risks are influenced by architectural features, with attention mechanisms (GAT) exacerbating leakage, while injective aggregation (GIN) provides comparatively stronger resilience.
We further evaluate several SOTA defense techniques, including differential privacy, gradient clipping, secure aggregation, model compression with quantization, and adversarial training.
We find that these techniques improve resilience only in specific settings and can also compromise model performance.
Overall, our work provides key insights toward privacy-preserving GNNs and highlights the need for more robust and efficient defenses.
We release our full methodology and artifacts.
\end{abstract}
\IEEEpeerreviewmaketitle


\section{Introduction}
\label{sec:introduction}

Graph neural networks (GNNs) \cite{zhang2019graph,karn2024benchmarking} are powerful tools for reasoning over problems with inherent graph structures.
For circuit design, GNNs help to model complex relationships among register-transfer level (RTL) or gate-level components, facilitating crucial tasks such as verification with high efficacy \cite{zhang2019circuit,kunal2023gnn,shahane2023graph}.
However,
the use of GNNs can also induce security risks \cite{zhang2024trustworthy,alrahis2023tt,guan2024graph}, with some overlooked so far.
For example, in gradient leakage attacks (GLAs)
	\cite{zhu2019deep,gong2023gradient,scheliga2023dropout}, attackers exploit the gradients computed during training to infer or reconstruct sensitive information from the data,
     potentially exposing information relevant to hardware security \cite{fang2024privacy}.

Here, we present the first comprehensive study of GLAs on GNNs in the context of circuit-design and hardware-security tasks, covering 
graph sample and aggregate (GraphSAGE) \cite{lu2023graphsage}, graph convolution network (GCN) \cite{zhang2019graph}, graph isomorphism network (GIN) \cite{alrahis2021omla}, and graph attention network (GAT) \cite{hu2023parsing}.
All GNNs are trained for gate classification \cite{alrahis2021omla} and detection of hardware Trojans (HTs) \cite{lashen2023trojansaint}, respectively, which serve as exemplary sensitive
applications.
We also benchmark several
SOTA defense techniques, namely differential privacy \cite{li2020defense}, gradient clipping \cite{wei2021gradient}, secure aggregation \cite{wang2024breaking}, model compression with quantization \cite{liu2021model},
and adversarial training \cite{wei2021gradient}.
We find that, while these
techniques provide some resilience, as in increasing the reconstruction errors imposed onto attackers, they are not sufficient to thwart GLAs and furthermore impact performance.


Our contributions are focused on quantifying this fundamental threat of GLAs on GNNs, forming the necessary first step for assessment of defense against end-to-end exploits.
That is, rather than limiting ourselves to specific downstream attacks, we establish that sensitive circuit data can be reconstructed in general from GNNs used for circuit design/analysis.
Our study not only confirms this threat, and the limitations of existing mitigation techniques, but also lays the groundwork for future attack/defense works.

In summary, our contributions are as follows:
\begin{enumerate}
\item We formulate and systematically evaluate gradient leakage attacks against GNNs in the context of circuit design and hardware-security applications.
\item We quantify leakage across four GNN architectures, two representative circuit-analysis tasks, and ISCAS'85/EPFL/TrustHub-derived benchmarks.
\item We analyze how leakage varies across gate types, Trojan labels, and GNN architectures.
\item We evaluate common mitigation mechanisms and show that their privacy--accuracy trade-offs are highly model- and task-dependent.
\end{enumerate}

We provide a full release at {\footnotesize \url{https://github.com/rkarn/GradientAttackGNNs}}~.

\begin{table*}[!t]
\footnotesize
\centering
\caption{Comparison of prior art and ours.}
\label{tab:prior_arts_comparison}
\setlength{\tabcolsep}{1.5pt}
\begin{tabular}{|p{0.4cm} |p{3.2cm}| p{3cm}| p{5cm}| p{6cm}|}
\hline
\textbf{Ref.} & \textbf{Models} & \textbf{Datasets} & \textbf{Metric rel\_l2} & \textbf{Scope} \\
\hline
\cite{zhu2019deep}     
& CNNs (LeNet, AlexNet)  
& MNIST, CIFAR-10  
& $\approx 0.01$ (MNIST)  
& Recovers training data by inverting gradients from CNNs.
\\
\hline

\cite{hitaj2017deep}   
& DNNs w/ GAN-based attack in collab.\ learning  
& MNIST, CIFAR-10  
& NA (visual reconstructions)  
& Exploits shared gradients to reconstruct sensitive inputs.
\\
\hline

\cite{geiping2020inverting} 
& CNNs in federat.\ learning  
& CIFAR-10, ImageNet subsets  
& $\approx 0.03$ (CIFAR-10), $\approx 0.05$ (ImageNet)  
& Attacks federated setups via gradient inversion.
\\
\hline

\cite{melis2019exploiting}      
& DNNs in collab.\ learning  
& CIFAR-10, IMDB
& NA (attribute inferences)  
& Shows unintended leakage through gradients.
\\
\hline

\cite{zhang2022survey}           
& Various CNNs 
& MNIST, CIFAR-10, ImageNet  
& Summarizes prior works ($\approx 0.01$–$0.05$)  
& Taxonomy of gradient inversion attacks, defenses.
\\
\hline

\cite{sinha2024gradient} 
& GNNs (GCN, GraphSAGE)  
& Cora, Citeseer, PubMed  
& $\approx 0.1$--$0.5$ (node features)  
& Gradient inversion attacks on GNNs for node- and graph-level tasks in generic graph datasets.
\\
\hline

Ours                            
& GNNs (GCN, GraphSAGE, GIN, GAT)  
& ISCAS’85, EPFL, TrustHub 
& $\approx 0.8$–$1.0$ (gate classification), $\approx 1.2$–$2.2$ (HT detection)  
& First comprehensive study of GLAs on circuit-trained GNNs for hardware-security tasks (gate classification, HT detection).
\\
\hline
\end{tabular}
\vspace{-10pt}
\end{table*}

\section{Background and Motivation}
\label{sec:preliminaries}

\subsection{Graph Neural Network (GNNs)}
\label{subsec:gcnn_preliminary}

Let $G=(V,E)$ represent an undirected graph with $N$ nodes ($V$) and edges ($E$).
Each node $v_i \in V$ has features $\mathbf{x}_i \in \mathbb{R}^{d}$, collectively forming $X \in \mathbb{R}^{N \times d}$.
The graph structure is encoded by an adjacency matrix $A \in \mathbb{R}^{N \times N}$ and degree matrix $D$ ($D_{ii} = \sum_j A_{ij}$) \cite{velivckovic2023everything,lachaud2022mathematical}. 
$\tilde{A} = A + I_{N}$ denote self-connections.
The hidden layer operates as:
\begin{equation}
    H^{(l+1)} = \sigma\left(\hat{A} H^{(l)} W^{(l)}\right)
    \label{eq:gcnn_layer}
\end{equation}
where
	$H^{(0)} = X$, $W^{(l)} \in \mathbb{R}^{d_l \times d_{l+1}}$ are learnable weights,
	$\hat{A} = D^{-\frac{1}{2}} \tilde{A} D^{-\frac{1}{2}}$ is the normalized adjacency matrix, and
	$\sigma$ is an activation function, e.g., ReLU.
This operation aggregates information similarly to image convolution \cite{han2022vision,hu2021graph}
and can also be viewed as message passing \cite{gama2020graphs}:
\begin{equation}
    h_i^{(l+1)} = \sigma\left( W^{(l)} \cdot \text{AGG}\left( \left\{ h_j^{(l)} \mid j \in \mathcal{N}(i) \cup \{i\} \right\} \right) \right)
    \label{eq:message_passing}
\end{equation}
where $\mathcal{N}(i)$ denotes node $i$'s neighbors and AGG aggregates features, typically via sum/mean/max operations \cite{zhang2019graph}.
The final layer $H^{(L)}$ uses softmax activation for node classification.

\subsection{Gradient Leakage Attacks (GLAs)}

GLAs \cite{zhu2019deep,gong2023gradient,scheliga2023dropout} refer to a class of malicious techniques where attackers exploit gradient information, collected during the training or inference process of an ML model, to reconstruct sensitive details of the underlying input data.
GLAs have been studied extensively in domains like image recognition, e.g., showing that visually similar images can be recovered from leaked gradients \cite{zhu2019deep}. More recently, gradient leakage has also been explored for graph neural networks in generic learning settings. For example, \cite{sinha2024gradient} study gradient inversion attacks on GNNs for node- and graph-classification tasks, while related works investigate reconstruction of graph structures and node features from gradients in federated learning settings. In contrast to these efforts, our work focuses on circuit-trained GNNs and hardware-security applications, including gate classification and hardware Trojan detection, and studies the interplay between GNN architectures, circuit-specific features, and defense mechanisms. Table~\ref{tab:prior_arts_comparison} summarizes prior art and ours.



\subsubsection{\uline{GLAs on GNNs for Circuit Design -- An Overlooked Threat}}
\label{subsec:grad_leakage_prelim}
For circuit design, such attacks have been overlooked so far.
To understand the motivation, an adversary might utilize GLAs on GNNs trained for circuit design and analysis to:

\ul{Reconstruct Sensitive Inputs.}
Recover design details embedded in the training set
\cite{zhu2019deep,fang2024privacy}.
For example, attackers may infer structural properties of security features like supervisor-mode logic,
    which may assist the analysis of such features in downstream attack scenarios.
    
\ul{Perform Membership Inference.}
Determine whether specific designs were used during training
\cite{hu2022membership}.
    This may provide information that could assist the analysis of logic locking schemes \cite{kamali2022advances} by identifying whether particularly vulnerable circuits or locking schemes were used for training 
    \cite{alrahis2021omla}.

\ul{Execute Model Inversion.}
Extract representative features or typical instances
\cite{song2022survey}.
For example, attackers could
identify patterns associated with false negatives for HT detection \cite{kunal2023gnn}, which may inform the design of evasive structures to evade detection \cite{bhattacharyay2022automated}.
For another example, by identifying true positives for security assessment on logic locking, adversaries can learn which gate structures are most promising to attack.

\ul{Extract Model Parameters.}
Infer model parameters and/or reverse-engineering of proprietary ML algorithms \cite{torrance2009state, wei2021gradient}.
Attackers can reconstruct the learned feature space of GNNs used for security assessment, gaining adversarial insights for, e.g., bypassing detection of counterfeiting or tampering \cite{meade2016netlist}.

These motivational examples underscore the need for robust assessment of GLAs on GNNs applied to sensitive circuits.


\label{subsec:leakage_attack}

\subsubsection{\uline{Mathematical Formulation}}
\label{subsec:GLA_math}

Building upon the GNN formalism in Section~\ref{subsec:gcnn_preliminary}, we now define GLAs on GNNs.
Let $\mathcal{L}: \mathbb{R}^{N \times d} \rightarrow \mathbb{R}$ denote the loss function used for training, e.g.,
	 the cross-entropy loss for node classification.
Given the fixed parameters $\{W^{(l)}\}_{l=0}^{L-1}$ and the normalized adjacency matrix $\hat{A}$, the loss is computed as:
\[
\mathcal{L}(X, \hat{A}, \{W^{(l)}\}) = \mathcal{L}\big(H^{(L)}(X, \hat{A}, \{W^{(l)}\})\big)
\]
where $H^{(L)}$ is the network output defined in Equation~\eqref{eq:gcnn_layer}.

Now, suppose that adversaries have access to the gradients with respect to the model’s weights:
\[
G^{(l)} = \nabla_{W^{(l)}} \mathcal{L}(X, \hat{A}, W^{(l)}), \quad l=0,1,\ldots,L-1.
\]

The adversaries' goal is to recover all or parts of the original features $X$.
To do so, a dummy feature matrix $\tilde{X} \in \mathbb{R}^{N \times d}$ is introduced.
The reconstruction of the original features is then posed as the following optimization problem:
\begin{equation}
    \min_{\tilde{X}} \;
 \sum_{l=0}^{L-1} \left\| \nabla_{W^{(l)}} \mathcal{L}(\tilde{X}, \hat{A}, W^{(l)}) - G^{(l)} \right\|_F^2
    \label{eq:reconstruction_obj}
\end{equation}
where $\|\cdot\|_F$ denotes the Frobenius norm.

For targeted reconstruction of a specific node $v_i \in V$, we denote the true feature as $\mathbf{x}_i \in \mathbb{R}^{d}$ and the corresponding dummy feature as $\tilde{\mathbf{x}}_i$.
The localized reconstruction objective can be written as:
\begin{equation}
    \min_{\tilde{\mathbf{x}}_i \in \mathbb{R}^{d}} \; \left\|
 \nabla_{W^{(l)}} \mathcal{L}(\tilde{\mathbf{x}}_i, \hat{A}, W^{(l)}) - g_i^{(l)} \right\|_2^2
    \label{eq:local_reconstruction}
\end{equation}
where $g_i^{(l)}$ represents the contribution of node $v_i$ to the gradient $G^{(l)}$, and $\|\cdot\|_2$ is the Euclidean norm.

\subsubsection{\uline{Metrics}}
\label{subsec:GLA_metrics}
To quantify GLAs, various metrics help to measure how closely the reconstructed node features $\tilde{\mathbf{x}}_i$ approximate the true features $\mathbf{x}_i$ for each node $v_i
\in V$.

\ul{Absolute L2 Error ($\mathrm{abs\_l2}$).}  
The absolute Euclidean distance between the reconstructed and true features is defined as:
\begin{equation}
\label{equ:abs_l2}
    \mathrm{abs\_l2}_i = \|\tilde{\mathbf{x}}_i - \mathbf{x}_i\|_2.
\end{equation}
The metric's optimum is $0$, i.e., for perfect reconstruction.

\ul{Relative L2 Error ($\mathrm{rel\_l2}$).}  
Normalizing reconstruction errors relative to the magnitude of the true feature vector, we define:
\begin{equation}
    \label{equ:rel_l2}
    \mathrm{rel\_l2}_i = \frac{\|\tilde{\mathbf{x}}_i - \mathbf{x}_i\|_2}{\|\mathbf{x}_i\|_2 + \epsilon},
\end{equation}
where $\epsilon > 0$ is a small constant.
As before, reaching $0$ indicates that reconstructed features closely match the true features.

\ul{ Cosine Similarity ($\mathrm{cos\_sim}$).}  
To evaluate the directional alignment between reconstructed and true features, we define:
\begin{equation}
    \label{equ:cos_sim}
    \mathrm{cos\_sim}_i = \frac{\tilde{\mathbf{x}}_i \cdot \mathbf{x}_i}{\|\tilde{\mathbf{x}}_i\|_2 \, \|\mathbf{x}_i\|_2 + \epsilon}.
\end{equation}
This metric ranges between $-1$ and $1$, with $1$ representing perfect alignment, $0$ indicating orthogonality (no directional similarity), and $-1$ indicating inverse alignment.
High cosine similarity implies that the reconstructed feature preserves the orientation of the original feature, even if the magnitude differs.

Together, these three metrics
capture complementary aspects of reconstruction efforts: \(\mathrm{abs\_l2}\) measures absolute deviation, \(\mathrm{rel\_l2}\) normalizes the error relative to feature magnitude, and \(\mathrm{cos\_sim}\) evaluates directional similarity.
A successful GLA aims to minimize both \(\mathrm{abs\_l2}\) and \(\mathrm{rel\_l2}\) to $0$, while maximizing \(\mathrm{cos\_sim}\) to $1$.

These metrics are consistent with prior works on gradient inversion and leakage analysis \cite{zhu2019deep,geiping2020inverting,gong2023gradient}, where reconstruction quality is evaluated using L2-based errors and cosine similarity. In our context of circuit-trained GNNs, these metrics capture the extent to which structural node features, e.g., fan-in/out, centrality, or distance-based attributes (node features are described next in Section~\ref{sec:attack_methodology}), can be inferred from gradients.
Hence, they serve as established proxies for quantifying feature-level information leakage from learned representations.

\subsubsection{\uline{Threat Model}}

For an attacker to exploit gradient leakage, they must gain access to the gradients computed during training or inference. In this work, we consider a standard gradient leakage setting commonly adopted in prior literature \cite{zhu2019deep,gong2023gradient}, where the attacker has access to model parameters and corresponding gradients (e.g., for a given training step or target node), and aims to reconstruct the underlying input features.
This setting reflects a white-box or honest-but-curious scenario and serves as a canonical baseline to assess the inherent leakage of a model.
This can be achieved through several means:
\begin{itemize}
    \item {Model Extraction}: Attackers may gain access to pre-trained model weights post-deployment \cite{wen2023adaptivenet}, allowing them to
    compute gradients on select inputs.
\item {Side-Channel Attacks}: Adversaries can monitor hardware-level computations during training, like memory-access patterns, power consumption, or timing variations, to infer gradient updates without direct model access \cite{chabanne2021side}.
\item {Federated Learning Scenarios}: If the model operates in a distributed training environment, gradients exchanged between clients and servers may be intercepted \cite{wei2021gradient}.
\item {Adversarial Querying}: Attackers interacting with a deployed model can carefully craft queries to infer responses that, when analyzed systematically \cite{li2020defense}, expose gradients.
\end{itemize}

Unlike direct data theft, these means and subsequent GLAs can expose sensitive features in a stealthy manner, making them difficult to detect and reinforcing the need for robust defenses.

\begin{figure*}[tb]
\centering
		\includegraphics[scale=0.85, trim = {0.5cm 0 0cm 0}, clip]
{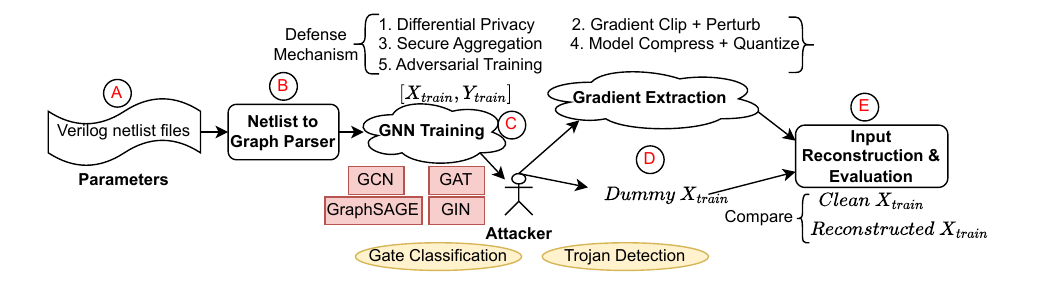} 
		\caption{Methodology for security evaluation of GNNs against GLAs.
		}
		\label{fig:Gradient_Attack_end_to_end}
\end{figure*}

\subsubsection{\uline{Mitigation Techniques}}
\label{subsec:prelim_prevention_measures}

As explained, GLAs exploit the information contained in gradient updates to reconstruct sensitive input data.
To mitigate such vulnerabilities, researchers have proposed several strategies that aim to obscure or limit
the information that gradients reveal.
However, as we later show, doing so is challenging for circuit-design tasks.

\ul{Differential Privacy.}
By injecting carefully calibrated noise into gradient updates, the training procedure ensures that the contribution of any individual data point is masked
\cite{li2020defense}.
This technique helps to prevent precise reconstruction of the original input while preserving general model performance\footnote{In this work, differential privacy is implemented via gradient clipping and noise injection following prior practice \cite{abadi2016deep}, where clipping bounds per-sample gradients and noise is added to obscure individual contributions.}

\ul{Gradient Clipping and Perturbation.}
Clipping restricts the magnitude of gradients, preventing large-scale updates that may leak more information
\cite{wei2021gradient}.
Along with perturbations, this approach further obfuscates the training process.

\ul{Secure Aggregation Protocols.}
In distributed/federated learning settings, secure aggregation allows the server to receive only aggregated gradients from multiple participants
\cite{wang2024breaking}.
By ensuring that individual gradient updates are never available in isolation, these protocols make it significantly more challenging for an adversary to reconstruct any specific data sample.

\ul{Model Compression and Quantization.}
Such techniques
reduce the precision of gradient updates
\cite{liu2021model}.
Lowering the resolution of the gradient information can inherently limit the granularity of data that is leaked, though this may come at cost for model accuracy.

\ul{Adversarial Training.}
Integrating adversarial examples during training not only makes models more robust to input perturbations but also helps to obfuscate gradient signals
\cite{wei2021gradient}.
This approach deliberately alters learning dynamics so that extracted gradients become less indicative of original input features.


\section{Methodology}
\label{sec:attack_methodology}

We devise an end-to-end methodology for security evaluation of GNNs used for circuit design and analysis against GLAs.
The workflow is outlined in Fig.~\ref{fig:Gradient_Attack_end_to_end} and explained next.

\textbf{\encircle{\textcolor{red}{A}} Input Processing.}
\label{subsec:verilog_netlists}
The workflow begins with processing the circuit netlists, along with parameters for training etc.

\textbf{\encircle{\textcolor{red}{B}} Graph Construction and Feature Setup for Sensitive Applications.}
\label{subsec:netlist2graph}
%
Each netlist
is converted into a graph where nodes represent gates and edges represent wires.
While doing so, relevant features for the sensitive circuit design/analysis task of interest are employed.
As motivated in Section~\ref{subsec:grad_leakage_prelim}, adversarial techniques like membership inference or model inversion can be employed against GNNs devised for security assessment of, e.g., logic locking and HT
detection.
Accordingly, we consider classification of gate types as a representative proxy task relevant to logic locking analysis, and binary classification of malicious vs. benign gates as a representative task for guiding adversarial efforts to bypass HT detection.

Following \cite{alrahis2021omla,lashen2023trojansaint} and the wider literature, we use the following node features:
\textit{fan\_in}, \textit{fan\_out}, \textit{dist\_to\_output}, \textit{is\_primary\_input}, \textit{is\_primary\_output}, \textit{is\_internal},
\textit{degree\_centrality}, \textit{betweenness\_centrality}, \textit{closeness\_centrality}, \textit{clustering\_coefficient}, \textit{avg\_fan\_in\_neighbors}, and \textit{avg\_fan\_out\_neighbors}.
The gate type and HT assignment, respectively,
are encoded as numerical and binary labels.



\textbf{\encircle{\textcolor{red}{C}} GNN Training with GLA Mitigations.}
\label{subsec:gcnn_training}
%
Training is performed using standard optimization techniques like gradient descent;
a detailed description and implementation can be found in our release.
At this stage, we also employ the GLA mitigation techniques outlined in Section~\ref{subsec:prelim_prevention_measures}.


\textbf{\encircle{\textcolor{red}{D}} Gradient Extraction.}
\label{subsec:gradient_extraction}
%
Here, we mimic attackers that extract gradients with respect to the model's parameters based on the loss observed at a targeted
node during training.
Algorithm~\ref{alg:GE} describes the extraction process.
In brief,\footnote{%
In more detail,
line~\ref{alg:GE:line1} initializes the clean feature matrix and line~\ref{alg:GE:line2} computes the output for the given graph.
The target node's output is isolated in line~\ref{alg:GE:line3}, after which the
   target loss is computed in line~\ref{alg:GE:line4}.
Gradients are cleared in line~\ref{alg:GE:line5}, before performing backpropagation in line~\ref{alg:GE:line6}.
Finally, gradients per layer are collected
   (line~\ref{alg:GE:line7}), the true target feature is recorded (line~\ref{alg:GE:line8}), and outputs are returned (line~\ref{alg:GE:line9}).
}
after the GNN computes the output for the entire graph, the loss is evaluated solely over the target node.
Then, gradients $\{G^{(l)}\}_{l=0}^{L-1}$ are obtained by backpropagation, and serve as leaked information for subsequent reconstruction.

\begin{algorithm}[!t]
\caption{Gradient Extraction}\label{alg:GE} 
\small
\SetAlgoLined
\KwIn{Graph $G=(V,E)$, clean feature matrix $X \in \mathbb{R}^{N \times d}$, trained GNN model with parameters $\{W^{(l)}\}_{l=0}^{L-1}$, target node index $i$, target label $y$}
\KwOut{Target gradients $\{G^{(l)}\}_{l=0}^{L-1}$, true target feature $\mathbf{x}_i^{\text{true}}$}
Initialize the clean feature matrix: $X_{\text{clean}} \gets X$\;
\label{alg:GE:line1}
Compute the model output: $\hat{Y} \gets \text{model}(G, X_{\text{clean}})$\;
\label{alg:GE:line2}
Extract the target node's output: $\hat{y}_i \gets \hat{Y}(i)$\;
\label{alg:GE:line3}
Compute the target loss: $\mathcal{L}_{\text{target}} \gets \mathcal{L}(\hat{y}_i, y)$\;
\label{alg:GE:line4}
Zero the model gradients\;
\label{alg:GE:line5}
Perform backpropagation to compute $\nabla_{W^{(l)}} \mathcal{L}_{\text{target}}$ for all layers $l$\;
\label{alg:GE:line6}
For $l = 0$ to $L-1$: \\
\hspace*{1cm} $G^{(l)} \gets \nabla_{W^{(l)}} \mathcal{L}_{\text{target}}$\;
\label{alg:GE:line7}
Set $\mathbf{x}_i^{\text{true}} \gets X_{\text{clean}}(i)$\;
\label{alg:GE:line8}
\Return $\{G^{(l)}\}_{l=0}^{L-1}$ and $\mathbf{x}_i^{\text{true}}$\;
\label{alg:GE:line9}
\end{algorithm}

\textbf{\encircle{\textcolor{red}{E}} Input Reconstruction and Evaluation.}
\label{subsec:input_reconstruction}
Next, the leaked gradients are used for input reconstruction as outlined in
Section~\ref{subsec:GLA_math}.
That is,
attackers iteratively update the dummy feature vector $\tilde{\mathbf{x}}_i$ until its corresponding gradients closely match the leaked gradients $\{G^{(l)}\}$.
The attackers' objective can also be defined as:
\[
\min_{\tilde{\mathbf{x}}_i} \; \sum_{l=0}^{L-1} \left\|
 \nabla_{W^{(l)}} \mathcal{L}(\tilde{\mathbf{x}}_i, y) - G^{(l)} \right\|_{2}^{2}\,,
\]
where $\mathcal{L}$ is the loss function and $\nabla_{W^{(l)}} \mathcal{L}(\tilde{\mathbf{x}}_i, y)$ denotes the gradients computed using the dummy feature.

The iterative procedure for updating $\tilde{\mathbf{x}}_i$ by attackers is below:
\begin{enumerate}
    \item {Initialization:} Set $\tilde{\mathbf{x}}_i$ to a random vector.
This is analogous to an initial guess or random noise.
\item {Comparison:} At each iteration, compute the gradients for $\tilde{\mathbf{x}}_i$ and measure the delta to the extracted gradients $\{G^{(l)}\}$.
\item {Optimization:} Update $\tilde{\mathbf{x}}_i$, using gradient descent, to minimize the objective function.
This step ensures that $\tilde{\mathbf{x}}_i$ converges to a value that reproduces the leaked gradients.
\end{enumerate}


Finally,
we compare the reconstructed $\tilde{\mathbf{x}}_i$ with the true $\mathbf{x}_i^{\text{true}}$ from Algorithm~\ref{alg:GE} (Line~\ref{alg:GE:line8}).
Using the metrics defined in Section~\ref{subsec:GLA_metrics}, such comparison serves as independent security evaluation of GNNs against GLAs.

\section{Experiments}
\label{sec:experiments}

\subsection{Setup}
\label{subsec:expt_testbench}

\subsubsection{\uline{Implementation}}
All experiments were implemented in Python, via Jupyter notebooks.
For GNN models, we utilize the Deep Graph Library (DGL) \cite{wang2019deep}.
For a case study on MNIST \cite{baldominos2019survey}, which serves to contextualize the impact of GLAs on traditional image classification and simpler models,
we use TensorFlow to devise a fully connected neural network (FCNN).
We implement the proposed methodology (Section~\ref{sec:attack_methodology}) in full, including mitigation techniques.
For the latter, we follow the default setups reported in their papers and codes.
GNN and GLA hyper-parameters are provided in Table~\ref{tab:gcnn_hyperparams}; further details and 
source codes are available at {\footnotesize \url{https://github.com/rkarn/GradientAttackGNNs}}~.

\begin{table}[!t]
\footnotesize
\centering
\vspace{10pt}
\caption{Hyper-parameters for GNNs and GLAs.}
\label{tab:gcnn_hyperparams}
\begin{tabular}{lcc}
\hline
\textbf{Parameter} & \textbf{Gate Classification} & \textbf{HT Detection} \\ \hline
Input Feature Dimension ($d$) & 13 & 26 \\
Hidden Layer Units & 32 & 96 \\
Number 
of Output Classes ($C$) & 8 & 2 \\
Learning Rate & 0.01 & 0.01 \\
Training Epochs & 50 & 50 \\
Loss Function & Cross-Entropy & Cross-Entropy \\
Optimizer & Adam & Adam \\
\hline
Attack Iterations & 300 & 300 \\
Dummy Optimizer Learning Rate & 0.1 & 0.1 \\
Regularization Weight (L2) & 0.001 & 0.001 \\ \hline
\end{tabular}
\vspace{-10pt}
\end{table}

\subsubsection{\uline{Datasets and Sensitive Applications}}
We utilize ISCAS'85 \cite{hansen1999unveiling} and EPFL \cite{amaru2015epfl} benchmarks, obtained via 
	{\footnotesize \url{https://github.com/jpsety/verilog_benchmark_circuits}}~.
{We perform node-level splits for each constructed graph, with $80/20\%$ train/test ratio.}

As indicated, we focus on logic locking and HT detection as sensitive GNN applications.
For logic locking, we consider gate classification.
With this generic approach, we assess the fundamental threat of GLAs without being limited to specific locking techniques.
For HT detection, likewise, we consider detection of distinctive gate structures.
For dataset preparation, we inject selected HT templates from the TrustHub suite, namely
\emph{Countermux}, \emph{FSMor}, and \emph{Andxor}, into the ISCAS'85 and EPFL netlists.
Each HT is implemented for rare trigger conditions. All HTs are included in our release.
\subsection{Results for Baseline GNNs}

First, we study the performance and GLA vulnerabilities for baseline models, i.e., without any mitigations in place.

\subsubsection{\uline{Performance}}
Results for all baseline GNNs for both sensitive tasks (and for the MNIST FCNN) are given in Table~\ref{tab:performance_metrics}.
Especially for GNNs, results are competitive with SOTA, confirming that our setup is practical and relevant.
Given the security focus of this work, we refrain from further performance comparisons with prior art.

\subsubsection{\uline{GLAs Overview}}
Recall that lower \textit{abs\_l2}/\textit{rel\_l2} and higher \textit{cos\_sim} values correspond to stronger leakage and better reconstruction of sensitive data by attackers
(Section~\ref{subsec:GLA_metrics}).
Also recall that \textit{cos\_sim} serves as an intuitive indicator of leakage, as it directly captures the alignment between reconstructed and original feature vectors.

See the \textit{No Defense} scenario in Table~\ref{tab:median_reconstruction_comparison}.
Across GNNs and tasks, the success of GLAs varies notably, as discussed next.
To support aggregate analysis, Table~\ref{tab:gla_relative_summary} summarizes GNN-wise mean leakage and relative defense effects, while Table~\ref{tab:gla_classwise_summary} reports class-wise no-defense means across GNNs.

\subsubsection{\uline{GLAs on Gate Classification}}
On average across all GNNs (Table~\ref{tab:gla_classwise_summary}),  \textit{OR} gates are most vulnerable, with the highest \textit{cos\_sim} (0.643) and relatively low \textit{rel\_l2} (0.89).
\textit{NAND}, \textit{OUTPUT}, and \textit{XOR} gates also show high \textit{cos\_sim} values, but their cosine similarities remain slightly below \textit{OR}; their \textit{rel\_l2} values are comparable, with \textit{NAND} and \textit{XOR} higher and \textit{OUTPUT} slightly lower.
Conversely, \textit{NOT} gates yield the highest \textit{rel\_l2} (1.38) and the lowest \textit{cos\_sim} (0.295), indicating the strongest resilience among gate types.

These observations are consistent with the structural roles of the corresponding gates.
Gates with more distinctive connectivity and functional behavior can induce more distinguishable gradient patterns, while unary and frequently occurring gates such as \textit{NOT} provide less distinctive local signatures.
Thus, the gate-wise results indicate that leakage is not uniform across circuit components, but depends on the structural and functional characteristics captured by the GNN.

Regarding the different GNNs, on average across all gates (Table~\ref{tab:gla_relative_summary}), GAT is the most vulnerable architecture, with the highest \textit{cos\_sim} (0.715) and the lowest \textit{abs\_l2} (5.380) and \textit{rel\_l2} (0.795).
This aligns with the attention mechanism in GAT, which learns node-specific weighting and can produce more distinguishable gradients.
GIN is the most resilient architecture, with the lowest \textit{cos\_sim} (0.193) and the highest L2 errors (\textit{abs\_l2} 7.061, \textit{rel\_l2} 1.405).
This is consistent with its injective aggregation and non-linear transformations, which reduce the direct interpretability of gradients.
GraphSAGE demonstrates intermediate leakage behavior, consistent with its neighborhood sampling and aggregation mechanism, which preserves local structural information but does not expose it as strongly as GAT in this task.

\begin{table}[!t]
\scriptsize
\centering
\vspace{15pt}
\caption{Baseline model performances.}
\label{tab:performance_metrics}
\renewcommand{\arraystretch}{1.15}
\setlength{\tabcolsep}{2.7pt}
\begin{tabular}{l|cccc|cccc|c}
\hline
\textbf{Metric} & 
\multicolumn{4}{c|}{\textbf{Gate Classification}} & 
\multicolumn{4}{c|}{\textbf{HT Detection}} &
\multicolumn{1}{c}{\textbf{MNIST}}
\\ \cline{2-10}
\textbf{[\%]} & \textbf{GCN} & \textbf{Gr.SAGE} & \textbf{GIN} & \textbf{GAT} & 
\textbf{GCN} & \textbf{Gr.SAGE} & \textbf{GIN} & \textbf{GAT} & \textbf{FCNN} \\ \hline
Train acc & 92.45  & 93.58 & 93.44 & 93.48 & 99.98 & 99.99 & 99.95 & 99.82 & 98.01 \\
Test acc & 92.43 & 93.42 & 93.31 & 93.08 & 99.96 & 99.98 & 99.94 & 99.61 & 97.13 \\
Precision & 89.8  & 91.48 & 91.78 & 91.20 & 99.96 & 99.98 & 99.94 & 99.74 & 97.15 \\
Recall & 92.43 & 93.42 & 93.31 & 93.08 & 99.96 & 99.98 & 99.94 & 99.22 & 97.13 \\
F1 Score & 90.18 & 91.96 & 92.02 & 91.41 & 99.96 & 99.98 & 99.94 & 99.48 & 97.13 \\ \hline
\end{tabular}
\vspace{-10pt}
\end{table}

\begin{table*}[!t]
\scriptsize
\centering
\caption{Comparison of GLA results across GNNs, tasks, and defenses. Aggregate GNN-wise averages and relative defense effects are summarized in Table~\ref{tab:gla_relative_summary}, while class-wise no-defense averages are summarized in Table~\ref{tab:gla_classwise_summary}.}
\label{tab:median_reconstruction_comparison}
\begin{tabular}{|p{0.5cm}|p{0.8cm}
                |p{0.35cm}p{0.35cm}p{0.65cm}
                |p{0.35cm}p{0.35cm}p{0.65cm}
                |p{0.35cm}p{0.35cm}p{0.65cm}
                |p{0.35cm}p{0.35cm}p{0.65cm}
                |p{0.35cm}p{0.35cm}p{0.35cm}
               |p{0.35cm}p{0.35cm}p{0.65cm}|}
\hline
\rowcolor{Gainsboro} \multicolumn{20}{|c|}{\textbf{Gate Classification}} \\ 
\hline 
\textbf{GNNs} & \textbf{Class} 
& \multicolumn{3}{c|}{\textbf{No Defense}} 
& \multicolumn{3}{c|}{\textbf{Diff.  Priv. \cite{li2020defense}}} 
& \multicolumn{3}{c|}{\textbf{Grad. Clip. \cite{wei2021gradient}}} 
& \multicolumn{3}{c|}{\textbf{Secure Aggr. \cite{wang2024breaking}}} 
& \multicolumn{3}{c|}{\textbf{Compr. \& Quant. \cite{liu2021model}}} 
& \multicolumn{3}{c|}{\textbf{Adv.  Train. \cite{wei2021gradient}}} \\
\cdashline{3-20}[0.4pt/2pt]
& \textbf{Type} 
& abs\_l2 & rel\_l2 & cos\_sim 
& abs\_l2 & rel\_l2 & cos\_sim 
& abs\_l2 & rel\_l2 & cos\_sim 
& abs\_l2 & rel\_l2 & cos\_sim
& abs\_l2 & rel\_l2 & cos\_sim
& abs\_l2 & rel\_l2 & cos\_sim \\
\hline
\multirow{8}{*}{GCN} 
  & AND    & 2.28 & 1.06 & 0.49 & 2.27 & 1.07 & 0.46 & 2.41 & 1.14 & 0.41 & 2.50 & 1.12 & 0.33 & 2.39 & 1.17 & 0.43 & 2.54 & 1.14 & 0.39 \\
  & INPUT  & 5.26 & 0.77 & 0.68 & 5.58 & 0.81 & 0.61 & 5.43 & 0.79 & 0.65 & 5.55 & 0.81 & 0.60 & 5.43 & 0.79 & 0.63 & 5.38 & 0.78 & 0.66 \\
  & NAND   & 1.81 & 0.77 & 0.73 & 2.22 & 0.94 & 0.60 & 2.30 & 1.01 & 0.60 & 2.70 & 1.18 & 0.40 & 1.70 & 0.60 & 0.79 & 1.87 & 0.75 & 0.70 \\
  & NOR    & 3.04 & 0.74 & 0.60 & 3.05 & 0.74 & 0.62 & 3.35 & 0.81 & 0.61 & 3.54 & 0.89 & 0.46 & 3.01 & 0.73 & 0.66 & 3.03 & 0.74 & 0.66 \\
  & NOT    & 2.43 & 1.46 & 0.26 & 2.29 & 1.25 & 0.34 & 1.89 & 1.06 & 0.49 & 2.32 & 1.34 & 0.29 & 2.10 & 1.18 & 0.39 & 2.13 & 1.13 & 0.33 \\
  & OR     & 1.45 & 0.65 & 0.76 & 1.34 & 0.57 & 0.81 & 1.36 & 0.50 & 0.87 & 2.03 & 0.92 & 0.55 & 1.57 & 0.64 & 0.77 & 1.21 & 0.52 & 0.87 \\
  & OUTPUT & 6.27 & 0.75 & 0.76 & 6.54 & 0.78 & 0.63 & 6.42 & 0.76 & 0.73 & 6.41 & 0.76 & 0.69 & 6.27 & 0.75 & 0.74 & 6.14 & 0.73 & 0.77 \\
  & XOR    & 22.06 & 0.94 & 0.74 & 22.16 & 0.94 & 0.77 & 22.62 & 0.96 & 0.62 & 22.43 & 0.96 & 0.68 & 22.39 & 0.95 & 0.61 & 22.29 & 0.95 & 0.69 \\
\hline
\multirow{8}{*}{\makecell{Graph\\SAGE}} 
  & AND    & 2.58 & 1.24 & 0.30 & 2.48 & 1.12 & 0.37 & 2.57 & 1.27 & 0.34 & 3.28 & 1.70 & 0.14 & 2.72 & 1.27 & 0.34 & 2.52 & 1.24 & 0.32 \\
  & INPUT  & 6.02 & 0.88 & 0.56 & 5.86 & 0.86 & 0.60 & 6.23 & 0.91 & 0.44 & 6.06 & 0.88 & 0.54 & 6.24 & 0.91 & 0.45 & 5.98 & 0.87 & 0.53 \\
  & NAND   & 2.01 & 0.86 & 0.66 & 2.36 & 1.02 & 0.53 & 2.37 & 1.04 & 0.64 & 2.84 & 1.24 & 0.42 & 1.70 & 0.71 & 0.75 & 1.85 & 0.80 & 0.72 \\
  & NOR    & 3.43 & 0.84 & 0.50 & 3.32 & 0.83 & 0.52 & 3.56 & 0.88 & 0.51 & 3.71 & 0.93 & 0.44 & 3.41 & 0.89 & 0.51 & 3.39 & 0.83 & 0.52 \\
  & NOT    & 2.74 & 1.51 & 0.21 & 2.71 & 1.57 & 0.29 & 2.62 & 1.51 & 0.12 & 2.28 & 1.30 & 0.26 & 2.99 & 1.79 & 0.08 & 2.76 & 1.53 & 0.16 \\
  & OR     & 1.01 & 0.41 & 0.90 & 1.17 & 0.49 & 0.82 & 1.29 & 0.53 & 0.86 & 1.70 & 0.75 & 0.65 & 1.27 & 0.51 & 0.82 & 0.99 & 0.44 & 0.88 \\
  & OUTPUT & 7.09 & 0.84 & 0.68 & 7.38 & 0.88 & 0.53 & 7.32 & 0.87 & 0.59 & 7.36 & 0.88 & 0.55 & 7.24 & 0.86 & 0.58 & 7.29 & 0.87 & 0.58 \\
  & XOR    & 21.93 & 0.93 & 0.63 & 22.41 & 0.95 & 0.51 & 22.21 & 0.95 & 0.54 & 22.32 & 0.95 & 0.54 & 22.02 & 0.94 & 0.62 & 22.03 & 0.94 & 0.62 \\
\hline
\multirow{8}{*}{GIN} 
  & AND    & 3.50 & 1.77 & 0.10 & 3.72 & 1.97 & 0.07 & 3.68 & 1.92 & 0.13 & 3.43 & 1.83 & 0.18 & 2.93 & 1.39 & 0.23 & 3.79 & 1.95 & 0.10 \\
  & INPUT  & 6.87 & 1.00 & 0.24 & 6.89 & 1.01 & 0.24 & 6.51 & 0.95 & 0.36 & 6.64 & 0.97 & 0.29 & 7.00 & 1.02 & 0.20 & 6.76 & 0.98 & 0.28 \\
  & NAND   & 3.71 & 1.74 & 0.29 & 3.91 & 2.04 & 0.16 & 4.29 & 2.12 & 0.09 & 4.11 & 2.02 & 0.09 & 4.18 & 2.03 & 0.06 & 4.15 & 2.02 & 0.04 \\
  & NOR    & 4.85 & 1.29 & 0.09 & 4.52 & 1.20 & 0.26 & 5.15 & 1.36 & 0.03 & 5.11 & 1.36 & 0.10 & 4.55 & 1.23 & 0.25 & 4.88 & 1.30 & 0.12 \\
  & NOT    & 2.91 & 1.64 & 0.23 & 3.28 & 1.90 & 0.07 & 3.47 & 2.08 & 0.13 & 3.27 & 1.99 & 0.20 & 3.54 & 2.14 & 0.14 & 2.92 & 1.75 & 0.28 \\
  & OR     & 3.67 & 1.86 & 0.10 & 3.48 & 1.83 & 0.17 & 4.19 & 2.11 & -0.07 & 3.67 & 1.85 & 0.21 & 3.73 & 1.95 & 0.05 & 3.27 & 1.65 & 0.19 \\
  & OUTPUT & 8.17 & 0.97 & 0.22 & 7.69 & 0.91 & 0.40 & 8.09 & 0.96 & 0.30 & 8.41 & 1.00 & 0.18 & 8.25 & 0.98 & 0.23 & 8.41 & 1.00 & 0.17 \\
  & XOR    & 22.81 & 0.97 & 0.27 & 22.72 & 0.97 & 0.32 & 22.60 & 0.96 & 0.32 & 22.78 & 0.97 & 0.28 & 22.78 & 0.97 & 0.27 & 22.82 & 0.97 & 0.29 \\
\hline
\multirow{8}{*}{GAT} 
  & AND    & 1.93 & 0.90 & 0.58 & 2.21 & 1.06 & 0.51 & 2.16 & 1.02 & 0.45 & 2.36 & 1.08 & 0.38 & 2.08 & 1.02 & 0.53 & 2.21 & 1.11 & 0.45 \\
  & INPUT  & 5.56 & 0.81 & 0.71 & 5.52 & 0.81 & 0.70 & 5.64 & 0.82 & 0.65 & 5.66 & 0.83 & 0.65 & 5.73 & 0.84 & 0.65 & 5.60 & 0.82 & 0.68 \\
  & NAND   & 1.88 & 0.66 & 0.79 & 2.04 & 0.79 & 0.71 & 1.79 & 0.67 & 0.78 & 2.45 & 0.94 & 0.48 & 1.78 & 0.65 & 0.84 & 1.89 & 0.71 & 0.73 \\
  & NOR    & 3.24 & 0.79 & 0.61 & 3.16 & 0.77 & 0.69 & 3.05 & 0.73 & 0.68 & 3.54 & 0.92 & 0.53 & 3.16 & 0.76 & 0.64 & 3.16 & 0.77 & 0.65 \\
  & NOT    & 1.70 & 0.91 & 0.48 & 1.89 & 1.05 & 0.40 & 2.02 & 1.12 & 0.40 & 1.87 & 1.03 & 0.38 & 2.13 & 1.25 & 0.40 & 1.86 & 0.98 & 0.42 \\
  & OR     & 1.53 & 0.64 & 0.81 & 1.59 & 0.68 & 0.78 & 1.75 & 0.71 & 0.73 & 2.09 & 0.91 & 0.57 & 1.42 & 0.60 & 0.90 & 1.51 & 0.63 & 0.85 \\
  & OUTPUT & 6.36 & 0.76 & 0.86 & 6.41 & 0.76 & 0.83 & 6.38 & 0.76 & 0.85 & 6.34 & 0.75 & 0.85 & 6.34 & 0.75 & 0.85 & 6.39 & 0.76 & 0.85 \\
  & XOR    & 20.84 & 0.89 & 0.88 & 21.07 & 0.90 & 0.87 & 20.73 & 0.88 & 0.89 & 20.96 & 0.89 & 0.88 & 20.69 & 0.88 & 0.89 & 20.72 & 0.88 & 0.89 \\
\hline
\rowcolor{Gainsboro} \multicolumn{20}{|c|}{\textbf{HT Detection}} \\ 
\hline 
\multirow{2}{*}{GCN} 
  & Clean   & 7.59 & 1.57 & 0.17 & 7.16 & 1.38 & 0.22 & 7.79 & 1.46 & 0.14 & 8.09 & 1.49 & 0.10 & 7.74 & 1.50 & 0.11 & 7.79 & 1.41 & -0.06 \\
  & HT  & 6.48 & 2.22 & 0.13 & 6.73 & 2.26 & 0.04 & 6.91 & 2.24 & -0.04 & 6.49 & 2.12 & 0.05 & 6.59 & 2.08 & 0.00 & 6.70 & 2.22 & 0.11 \\
\hline
\multirow{2}{*}{\makecell{Graph \\SAGE}} 
  & Clean   & 3.34 & 0.85 & 0.50 & 4.69 & 1.41 & 0.22 & 3.34 & 1.02 & 0.38 & 5.57 & 1.30 & 0.19 & 4.24 & 1.34 & 0.34 & 2.82 & 0.83 & 0.53 \\
  & HT  & 4.95 & 0.49 & 0.89 & 1.02 & 0.40 & 0.94 & 0.99 & 0.41 & 0.91 & 8.63 & 2.57 & 0.17 & 1.88 & 0.69 & 0.87 & 1.40 & 0.45 & 0.92 \\
\hline
\multirow{2}{*}{GIN} 
  & Clean   & 4.66 & 1.20 & -0.05 & 4.33 & 1.24 & -0.02 & 3.93 & 1.19 & -0.02 & 5.55 & 1.36 & -0.10 & 3.99 & 1.12 & 0.02 & 4.24 & 1.23 & -0.06 \\
  & HT  & 8.07 & 1.55 & 0.15 & 4.69 & 1.41 & 0.17 & 3.56 & 1.32 & 0.26 & 5.42 & 1.37 & 0.18 & 4.78 & 1.42 & 0.19 & 3.77 & 1.22 & 0.21 \\
\hline
\multirow{2}{*}{GAT} 
  & Clean   & 3.8 & 0.9 & -0.05 & 4.2 & 1.0 & -0.03 & 3.5 & 0.85 & -0.02 & 5.0 & 1.2 & -0.08 & 4.0 & 0.95 & 0.01 & 3.2 & 0.8 & -0.04 \\
  & HT  & 6.5 & 1.4 & 0.18 & 5.0 & 1.3 & 0.17 & 5.8 & 1.35 & 0.22 & 7.2 & 1.45 & 0.16 & 5.5 & 1.35 & 0.19 & 4.5 & 1.1 & 0.25 \\
\hline
\end{tabular}
\end{table*}

\begin{table*}[!t]
\footnotesize
\centering
\vspace{10pt}
\caption{Mean GLA results and relative changes with respect to `no defense' calculated from Table~\ref{tab:median_reconstruction_comparison}. For GLAs under defense, values are percentage changes averaged over all classes. Positive $\Delta$rel\_l2 and negative $\Delta$cos\_sim indicate reduced leakage.}
\label{tab:gla_relative_summary}
\setlength{\tabcolsep}{3pt}
\renewcommand{\arraystretch}{1.1}
\begin{tabular}{|l|l|ccc|cc|cc|cc|cc|cc|}
\hline
\multirow{2}{*}{\textbf{Task}} &
\multirow{2}{*}{\textbf{GNN}} &
\multicolumn{3}{c|}{\textbf{No Defense Mean}} &
\multicolumn{2}{c|}{\textbf{Diff. Priv.}} &
\multicolumn{2}{c|}{\textbf{Grad. Clip.}} &
\multicolumn{2}{c|}{\textbf{Secure Aggr.}} &
\multicolumn{2}{c|}{\textbf{Compr. \& Quant.}} &
\multicolumn{2}{c|}{\textbf{Adv. Train.}} \\
\cline{3-15}
& & abs & rel & cos
& $\Delta$rel & $\Delta$cos
& $\Delta$rel & $\Delta$cos
& $\Delta$rel & $\Delta$cos
& $\Delta$rel & $\Delta$cos
& $\Delta$rel & $\Delta$cos \\
\hline
\multirow{4}{*}{Gate Type}
& GCN       & 5.575 & 0.893 & 0.628 & -0.6 & -3.6 & -1.5 & -0.8 & +11.8 & -20.3 & -4.6 &  0.0 & -5.6 & +1.0 \\
& GraphSAGE & 5.851 & 0.939 & 0.555 & +2.8 & -6.1 & +6.0 & -9.0 & +14.9 & -20.3 & +4.9 & -6.5 & +0.1 & -2.5 \\
& GIN       & 7.061 & 1.405 & 0.193 & +5.2 & +9.7 & +10.9 & -16.2 & +6.7 & -0.6 & +4.2 & -7.1 & +3.4 & -4.5 \\
& GAT       & 5.380 & 0.795 & 0.715 & +7.2 & -4.0 & +5.5 & -5.1 & +15.6 & -17.5 & +6.1 & -0.4 & +4.7 & -3.5 \\
\hline

\multirow{4}{*}{HT}
& GCN       & 7.035 & 1.895 & 0.150 & -4.0 & -13.3 & -2.4 & -66.7 & -4.7 & -50.0 & -5.5 & -63.3 & -4.2 & -83.3 \\
& GraphSAGE & 4.145 & 0.670 & 0.695 & +35.1 & -16.5 & +6.7 & -7.2 & +188.8 & -74.1 & +51.5 & -13.0 & -4.5 & +4.3 \\
& GIN       & 6.365 & 1.375 & 0.050 & -3.6 & +50.0 & -8.7 & +140.0 & -0.7 & -20.0 & -7.6 & +110.0 & -10.9 & +50.0 \\
& GAT       & 5.150 & 1.150 & 0.065 &  0.0 & +7.7 & -4.3 & +53.8 & +15.2 & -38.5 &  0.0 & +53.8 & -17.4 & +61.5 \\
\hline
\end{tabular}
\end{table*}

\begin{table}[!t]
\footnotesize
\centering
\caption{No-defense class-wise mean GLA results averaged across GNNs, calculated from Table~\ref{tab:median_reconstruction_comparison}.}
\label{tab:gla_classwise_summary}
\renewcommand{\arraystretch}{1.1}
\begin{tabular}{l|lccc}
\hline
\textbf{Task} & \textbf{Class} & \textbf{abs\_l2} & \textbf{rel\_l2} & \textbf{cos\_sim} \\
\hline
\multirow{8}{*}{Gate Type}
& AND    & 2.572 & 1.243 & 0.368 \\
& INPUT  & 5.927 & 0.865 & 0.548 \\
& NAND   & 2.352 & 1.008 & 0.618 \\
& NOR    & 3.640 & 0.915 & 0.450 \\
& NOT    & 2.445 & 1.380 & 0.295 \\
& OR     & 1.915 & 0.890 & 0.643 \\
& OUTPUT & 6.972 & 0.830 & 0.630 \\
& XOR    & 21.910 & 0.932 & 0.630 \\
\hline
\multirow{2}{*}{HT}
& Clean  & 4.848 & 1.130 & 0.143 \\
& Trojan & 6.500 & 1.415 & 0.338 \\
\hline
\end{tabular}
\vspace{-10pt}
\end{table}

\subsubsection{\uline{GLAs on HT Detection}}
On average across all GNNs (Table~\ref{tab:gla_classwise_summary}), the \textit{Trojan} class is more vulnerable, with \textit{cos\_sim} of 0.338, more than twice that of the \textit{Clean} class (0.143).
At the same time, the L2 errors are higher for the \textit{Trojan} class (\textit{abs\_l2} 6.5 vs.\ 4.848, \textit{rel\_l2} 1.415 vs.\ 1.13).
This indicates that GLAs recover stronger directional information for Trojan-related features, while the numerical reconstruction remains less accurate.
This trend is consistent with Trojan circuitry introducing distinctive structural patterns, such as trigger-related logic, that are reflected in the gradients.

Regarding the different GNNs, on average across both classes (Table~\ref{tab:gla_relative_summary}), GIN is again the most resilient architecture, with \textit{cos\_sim} of 0.05.
GAT comes second, with \textit{cos\_sim} of 0.065, whereas GraphSAGE is the most vulnerable, with the highest \textit{cos\_sim} of 0.695 and the lowest \textit{rel\_l2} of 0.67.
This differs from gate classification, showing that leakage behavior is task-dependent.
For HT detection, GraphSAGE's neighborhood aggregation captures localized structural patterns associated with Trojan insertion, which are reflected more strongly in the gradients.
GIN remains comparatively resilient, consistent with the gradient-obfuscating effect of its non-linear injective aggregation.
GAT shows lower leakage here than in gate classification, indicating that attention-based gradients are less exposed in this binary HT-detection setting than in the multi-class gate-classification setting.

\subsection{Results for GNNs with GLA Mitigations}
\label{sec:GNN_def}

Second, we study GLA risks and performance in depth for hardened models with different mitigations in place.
Test accuracies are given in Table~\ref{tab:gnn_defense_performance} and GLA results are given in Tables~\ref{tab:median_reconstruction_comparison} and \ref{tab:gla_relative_summary}.

\subsubsection{\uline{Performance for Gate Classification}}
For gate classification, several defenses slightly improve accuracy for selected models. In particular, GIN benefits from gradient clipping, secure aggregation, compression/quantization, and adversarial training, while GraphSAGE benefits from secure aggregation, reaching the largest gain of +2.15 percentage points  (ppts).
This indicates that these models were slightly underfitted initially, which is confirmed by Table~\ref{tab:performance_metrics}.
Both GAT and GIN are remarkably robust across most defenses, with -1.84 ppts reduction at worst,
whereas GraphSAGE suffers notably under differential privacy (-31.02 ppts) and gradient clipping (-15.35 ppts).
This can be explained by the respective GNN mechanisms: for GAT and GIN, noises induced by defenses can be largely `overlooked' in attention and injective aggregation mechanisms, whereas GraphSAGE's accumulation approach also accumulates noise, leading to larger errors.

\subsubsection{\uline{GLAs on Gate Classification}}
On average across all gates (Table~\ref{tab:gla_relative_summary}), secure aggregation is the most effective defense for GAT (\textit{cos\_sim} -17.5\%, \textit{rel\_l2} +15.6\%), GraphSAGE (\textit{cos\_sim} -20.3\%), and GCN (\textit{cos\_sim} -20.3\%).
For GIN, gradient clipping is most effective in reducing \textit{cos\_sim} (-16.2\%, with \textit{rel\_l2} +10.9\%).
Notably, differential privacy weakens the baseline resilience of GIN, increasing \textit{cos\_sim} by +9.7\%.
Regarding different gates, vulnerable combinations of gates and models do not benefit from defenses, e.g., \textit{cos\_sim} for \textit{OUTPUT} gates on GAT can at best (using differential privacy) be reduced by only
-3.5\%.
For inherently resilient gates like \textit{NOT}, some defenses are counterproductive on less resilient models, e.g., gradient clipping on GCN increases \textit{cos\_sim} from 0.26 to 0.49, whereas
some defenses are effective on resilient models, e.g., gradient clipping on GIN reduces \textit{cos\_sim} for \textit{NOT} gates from 0.23 to 0.13.
These diverse findings clearly show that defenses need to be carefully assessed before applying them to different GNN models for this sensitive task.

Despite notable improvements for selected defenses,
the final metrics still show significant leakage, suggesting that one cannot fully compensate for fundamental leakage arising from GNN architectures and their cores.
Also, the ranking of different GNNs remains similar:
GAT is still most vulnerable, with \textit{cos\_sim} values around 0.59--0.7 across defenses (vs.\ 0.715 without defenses),
and GIN is still most resilient with \textit{cos\_sim} as low as 0.161 (vs.\ 0.193 without defenses).

\subsubsection{\uline{Trade-Offs for Gate Classification}}
The performance cost induced by defenses are generally manageable, but security gains are also moderate.
Best trade-offs are achieved with secure aggregation for GAT, GCN, GraphSAGE, and with gradient clipping for GIN.

\begin{table}[!t]
\centering
\scriptsize
\vspace{10pt}
\caption{Hardened model performances: test accuracy [\%].}
\label{tab:gnn_defense_performance}
\setlength{\tabcolsep}{2.5pt}
\begin{tabular}{l|cccc|cccc}
\hline
\multirow{2}{*}{\textbf{Mechanism}} &
\multicolumn{4}{c|}{\textbf{Gate Classification}} &
\multicolumn{4}{c}{\textbf{HT Detection}} \\
\cline{2-9}
& \textbf{GCN} & \textbf{Gr.SAGE} & \textbf{GIN} & \textbf{GAT} 
& \textbf{GCN} & \textbf{Gr.SAGE} & \textbf{GIN} & \textbf{GAT} \\
\hline
No Defense & 92.43 & 93.42 & 93.31 & 93.08 & 99.96 & 99.98 & 99.94 & 99.61 \\
Diff. Priv.  & 89.23 &  62.40 & 92.04 & 91.68 & 53.59 &  42.63 & 24.95 & 82.24 \\
Grad. Clip. & 88.76 & 78.07 & 93.97  & 92.00 & 99.94 & 75.05  & 99.73 & 86.93 \\
Secure Aggr.  & 83.96 & 95.57 & 95.23 & 94.52 & 99.91 & 24.95 & 24.95 &  93.45\\
Compr. \& Quant. & 91.23  & 84.96 & 95.29 & 91.24 & 99.86 & 75.09 & 99.94 & 91.24 \\
Adv. Train. & 91.32 & 88.82 & 95.27 & 91.36 & 99.95 & 75.05 & 49.89 & 83.42 \\
\hline
\end{tabular}
\vspace{-10pt}
\end{table}

\begin{figure}[!t]
	\begin{center}
		\includegraphics[trim = {0.0cm 0.2cm 0cm 0.2cm}, clip, width=\columnwidth]{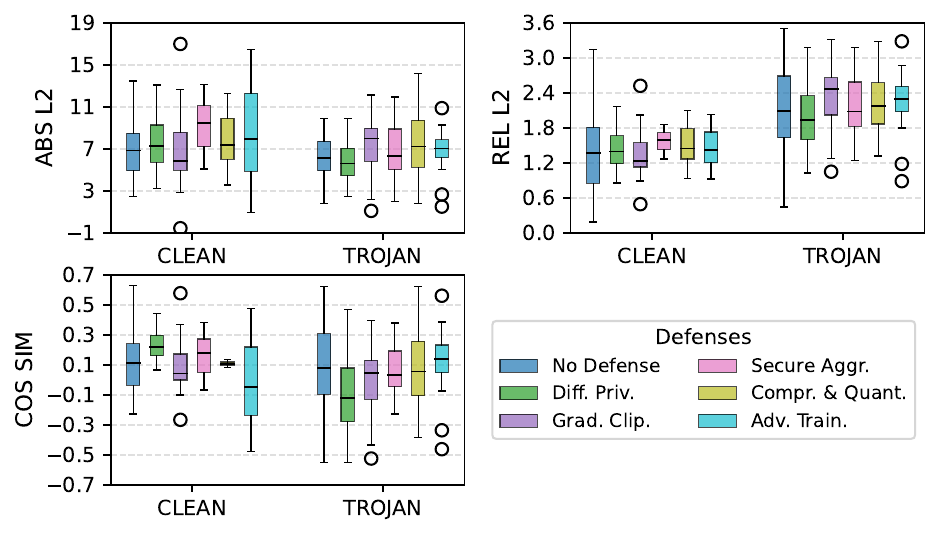} 
    \vspace{1pt}
        \caption{GLA results on GCN trained for HT detection.}
		\label{fig:boxplot_HT_alldefense}
	\end{center}
    \vspace{-10pt}
\end{figure}

\subsubsection{\uline{Performance for HT Detection}}
All defenses lead to significant performance drops for at least some GNNs, necessitating their careful assessment before application.
For example, secure aggregation undermines both GraphSAGE and GIN by approximately 75 percentage points, dropping their accuracies to 24.95\%.
Interestingly, GCN is the most robust architecture under most defenses, maintaining approximately 99.9\% accuracy except under differential privacy.
This suggests that its built-in, structural aggregation approach -- unlike secure aggregation applied on top of other models --
is highly stable.
In any case, recall that all defenses were applied with their default settings, possibly leaving room for further improvements.\footnote{%
The corresponding, significant effort to explore hyper-parameter tuning vs.\ accuracy vs.\ resilience, for all defenses and across all GNNs, can be scope for future work.}

\subsubsection{\uline{GLAs on HT Detection}}
As summarized in Table~\ref{tab:gla_relative_summary}, secure aggregation is by far the most effective defense for GraphSAGE (which showed the greatest baseline vulnerability), namely reducing \textit{cos\_sim} by -74.1\% and increasing \textit{rel\_l2} by +188.8\%,
respectively, on average.
Secure aggregation is also effective for GIN and GAT (\textit{cos\_sim} -20.0\% and -38.5\%).
For GCN, adversarial training is most effective (\textit{cos\_sim} -83.3\%), followed by gradient clipping and compression (\textit{cos\_sim} -66.7\% and -63.3\%).
Notably, multiple defenses can worsen the leakage across different GNNs, e.g.,
adversarial training on GraphSAGE increases \textit{cos\_sim} by +4.3\% or, representing the worst offender, gradient clipping on GIN increases \textit{cos\_sim} by +140\%.
Thus, even more than with gate classification, defenses need to be carefully assessed.

The ranking of GNNs shifts after mitigations are applied, indicating that one can partially compensate for GNN-inherent leakage for HT detection, which differs from gate classification.
For example, secure aggregation applied on GraphSAGE reduces its average \textit{cos\_sim} to approximately 0.18, substantially narrowing the gap to the more resilient models.
Even more significant is adversarial training applied on GCN, which ranked in the middle for baseline resilience, pushing its average \textit{cos\_sim} to a
near-zero value of 0.025. See also Fig.~\ref{fig:boxplot_HT_alldefense} for more detailed results for GCN, and note that further plots for other models and settings are provided in Appendix.

\subsubsection{\uline{Trade-Offs for HT Detection}}
For this task, trade-offs are more difficult to navigate.
For example,  both the performance and resilience of inherently resilient models can be undermined by some defenses, e.g., adversarial training on GIN.
Still, one `tower of strength' arises:
GCN almost always achieves high resilience with high performance.
For example, adversarial training on GCN is highly effective (reducing leakage by -83.3\%) while maintaining near-perfect accuracy (99.95\%).

\subsection{GLAs in Circuits vs. Image Domain}

\begin{table}[!t]
\centering
\footnotesize
\vspace{10pt}
\renewcommand{\arraystretch}{1.1}
\caption{GLA results on MNIST FCNN, for reference/context. This experiment serves as a sanity check for our implementation and as a baseline from prior gradient inversion literature.}
\label{tab:mnist_fcn_results}
\begin{tabular}{lccc}
\hline
\textbf{Defense Mechanism} & \textbf{Abs L2} & \textbf{Rel L2} & \textbf{Cosine Sim.} \\
\hline
None                              & 3.0350 & 0.3926 & 0.9209 \\
Differential Privacy              & 2.9191 & 0.3755 & 0.9270 \\
Gradient Clip 			& 8.7740 & 1.0425 & 0.7045 \\
Model Quantization  		& 3.1581 & 0.4061 & 0.9185 \\
Adversarial Training              & 2.3308 & 0.3027 & 0.9468 \\
Secure Aggregation                & 14.4771 & 1.6368 & 0.3406 \\
\hline
\end{tabular}
\vspace{-10pt}
\end{table}

To contextualize the vulnerability of circuit-trained GNNs against GLAs, we also report on an established GLA benchmark, namely MNIST image detection, while attacking a simpler FCNN, in Table~\ref{tab:mnist_fcn_results}.
We note low errors (\textit{abs\_l2} $\approx 2.3$--$14.5$, \textit{rel\_l2} $\approx 0.3$--$1.6$) with high \textit{cos\_sim} values ($>0.7$ in most cases), indicating strong gradient leakage.
In contrast, in the circuit domain, we saw generally higher \textit{abs\_l2} and \textit{rel\_l2} values and lower \textit{cos\_sim} values.

This can be attributed to the discrete and sensitive nature of circuit features.
For example, consider \textit{fan\_in}: even a small absolute difference 
can lead to a significant shift/error in the reconstruction of the underlying circuit structures, whereas in the image domain, a small difference between true and reconstructed pixel values would be largely imperceptible
(at least to humans).
Thus, while GLAs threaten both domains, the circuit domain exhibits some degree of inherent resilience.
Still, we must caution again here that this does not hold universally true: we found significant variability for leakage across GNN models, defenses, and sensitive tasks.

We include MNIST only as a sanity check for our implementation; all main conclusions in this work are based on circuit-trained GNNs.

\section{Conclusion and Future Work}
\label{sec:conclusion}

We have shown that circuit-tailored GNNs introduce fundamental leakage risks.
Our work provides multiple key findings.
First, GLAs can reveal sensitive input features across different GNN architectures and circuit-analysis tasks.
These risks should be evaluated in an architecture- and data-specific context.
For example, attention mechanisms (GAT) show high leakage in gate classification, whereas non-linear injective aggregation (GIN) provides stronger resilience across both studied tasks.
Second, the benefits of SOTA defense techniques are highly specific and require careful assessment.
While some techniques can significantly improve resilience, such as secure aggregation for GraphSAGE on HT detection, they do not consistently mitigate architecture-dependent leakage across tasks.
Third, SOTA defenses can even be detrimental, both to performance and resilience, for selected models such as GIN.
Consequently, future privacy-sensitive GNN settings for circuit design should prioritize robust backbones, carefully evaluate add-on defenses, and consider accuracy--privacy trade-offs.
To support future work in this direction, we release our evaluation methodology in full.

Future work includes extending this analysis to end-to-end attack scenarios and systematically exploring defense configurations through hyperparameter tuning.
Another promising direction is the development of inherently privacy-preserving GNN architectures tailored for circuit design and hardware-security applications.

\bibliographystyle{IEEEtran}
\bibliography{paper}

\appendix

\section{Defense Mechanisms for Gradient Leakage Attack}

\subsection{Differential Privacy}
\label{subsec:dp_defense}

To mitigate the risk of sensitive information being recovered via gradient inversion, we integrate a differential-privacy mechanism into the gradient extraction process, by perturbing the computed gradients with Gaussian noise. Let
\[
G^{(l)}_{\text{true}} = \nabla_{W^{(l)}} \mathcal{L}(\hat{Y}, Y)
\]
denote the true gradient of the loss function \(\mathcal{L}\) with respect to the model parameters \(W^{(l)}\) at layer \(l\), where \(\hat{Y}\) and \(Y\) are the model's prediction and the ground truth, respectively. We then make the gradient private by adding Gaussian noise:
\[
G^{(l)}_{\text{dp}} = G^{(l)}_{\text{true}} + \mathcal{N}(0, \sigma^2 I),
\]
where \(\mathcal{N}(0, \sigma^2 I)\) denotes Gaussian noise with zero mean and covariance \(\sigma^2 I\), and \(I\) is the identity matrix. In our experiments, we set \(\sigma = 0.1\), i.e., the privacy noise multiplier is 0.1.

Consequently, the reconstruction objective for the gradient leakage attack is modified to minimize the difference between the noisy gradients \(G^{(l)}_{\text{dp}}\) and the gradients computed from the dummy input \(\tilde{X}\), given by:
\[
\min_{\tilde{X}} \; \sum_{l=0}^{L-1} \left\| \nabla_{W^{(l)}} \mathcal{L}(\tilde{X}) - G^{(l)}_{\text{dp}} \right\|_F^2.
\]

In short, by injecting Gaussian noise with \(\sigma = 0.1\) into the gradients, we seek to obscure sensitive information while maintaining sufficient utility for model training and evaluation.

 \begin{figure}[!t]
	\begin{center}
		\includegraphics[scale=0.58, trim = {0cm 0cm 0cm 0cm}, clip]{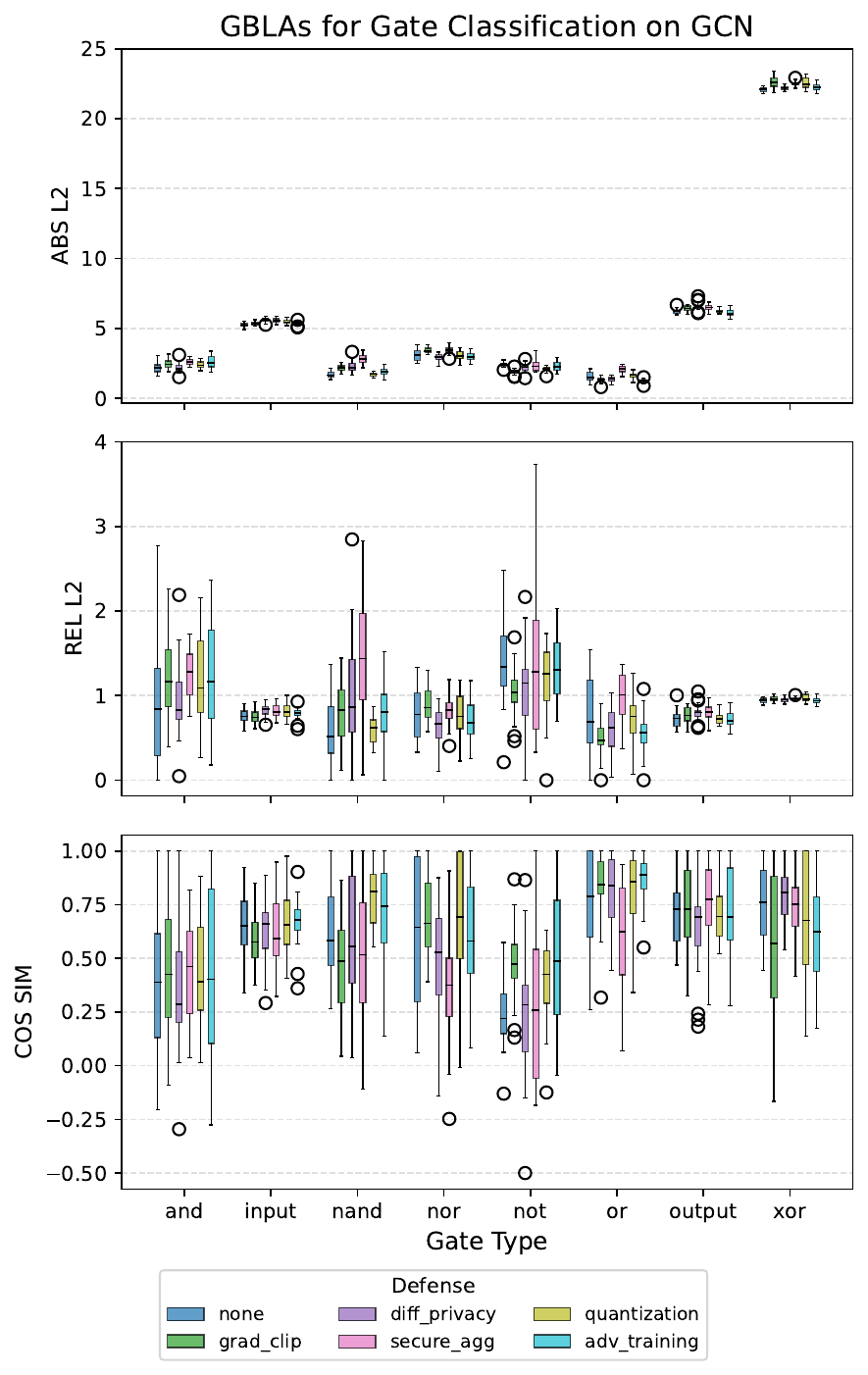} 
		\caption{\small Gate Classification: Reconstruction errors under different defense mechanism for GCN. } 
		\label{fig:gate_metrics_boxplots_combined_gcn}
	\end{center}
\end{figure}

 \begin{figure}[!t]
	\begin{center}
		\includegraphics[scale=0.58, trim = {0cm 0cm 0cm 0cm}, clip]{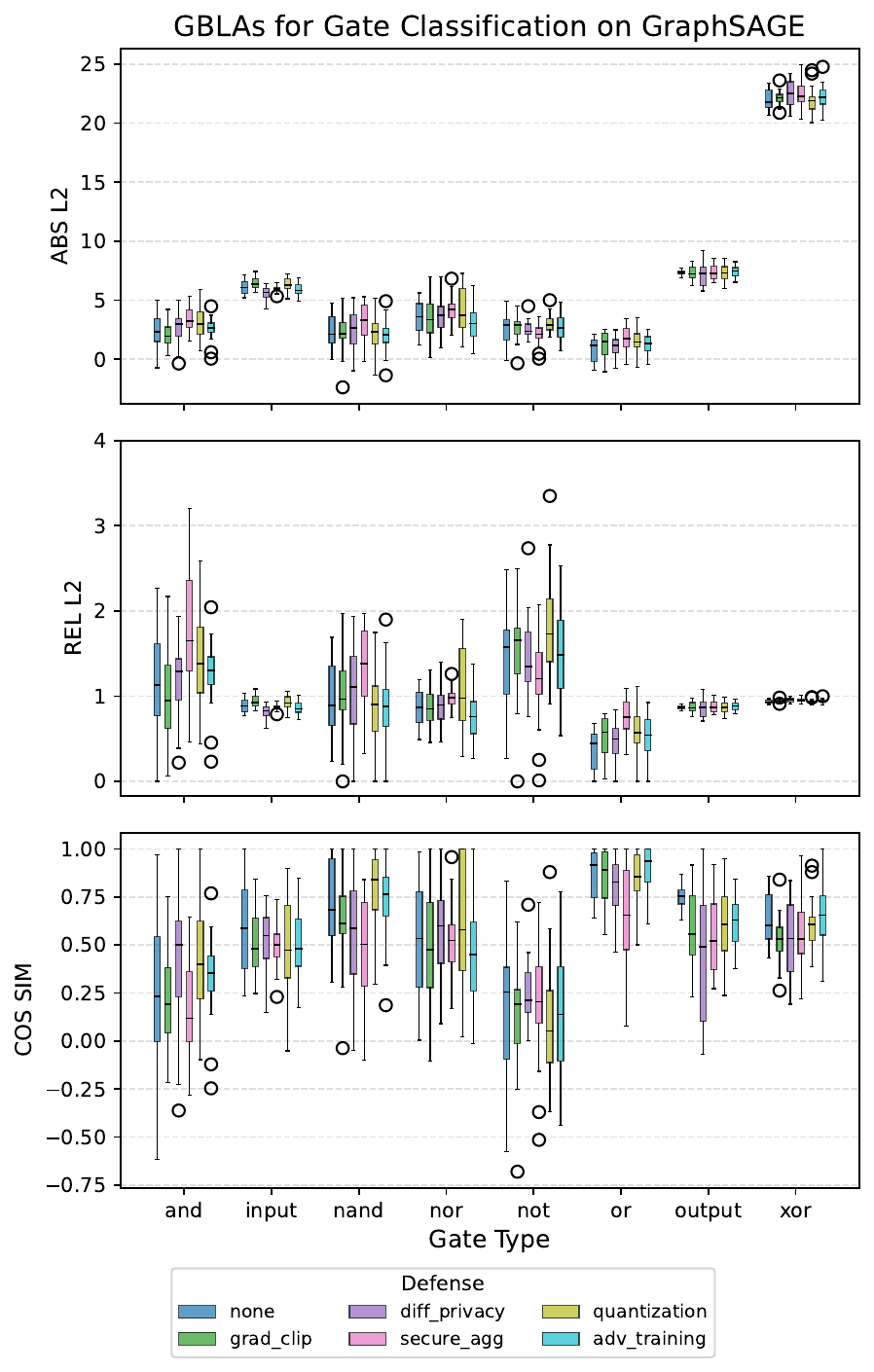} 
		\caption{\small Gate Classification: Reconstruction errors under different defense mechanism for GraphSAGE. } 
		\label{fig:gate_metrics_boxplots_combined_graphsage}
	\end{center}
\end{figure}

\begin{figure}[!t]
	\begin{center}
		\includegraphics[scale=0.58, trim = {0cm 0cm 0cm 0cm}, clip]{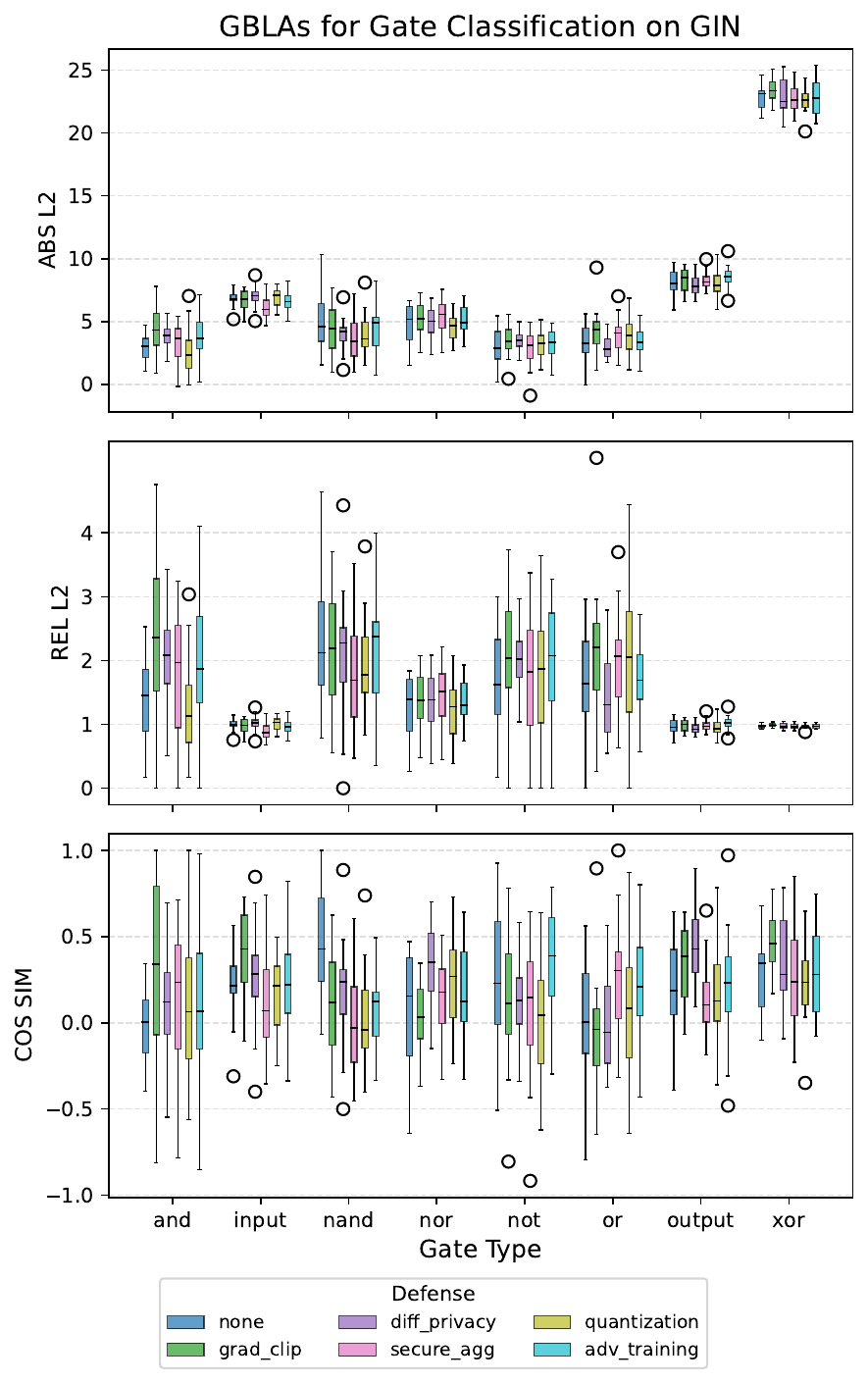} 
		\caption{\small Gate Classification: Reconstruction errors under different defense mechanism for GIN. } 
		\label{fig:gate_metrics_boxplots_combined_GIN}
	\end{center}
\end{figure}

\begin{figure}[!t]
	\begin{center}
		\includegraphics[scale=0.58, trim = {0cm 0cm 0cm 0cm}, clip]{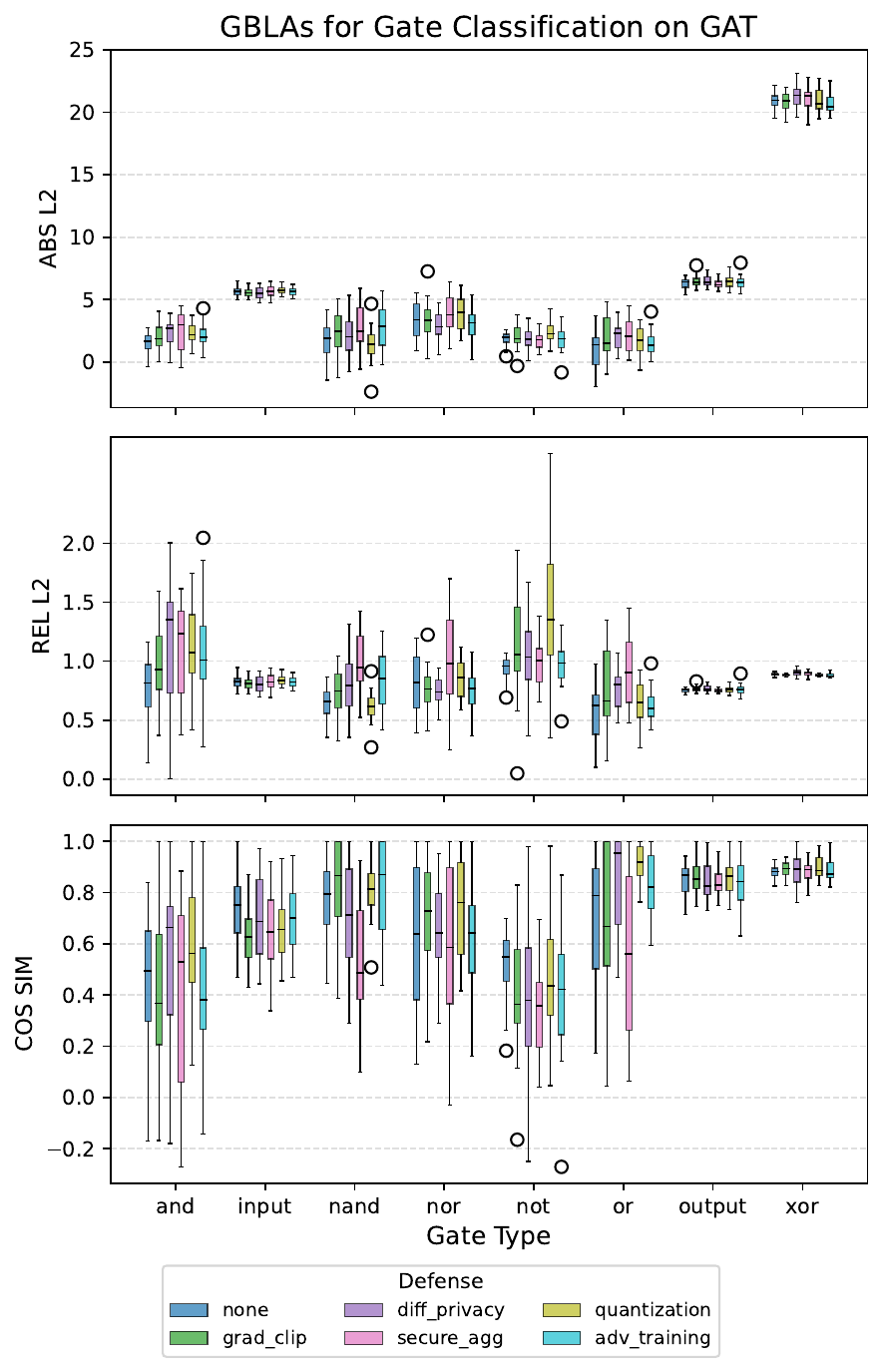} 
		\caption{\small Gate Classification: Reconstruction errors under different defense mechanism for GAT. } 
		\label{fig:gate_metrics_boxplots_combined_GAT}
	\end{center}
\end{figure}

\begin{figure}[!t]
	\begin{center}
		\includegraphics[scale=0.58, trim = {0cm 0cm 0cm 0cm}, clip]{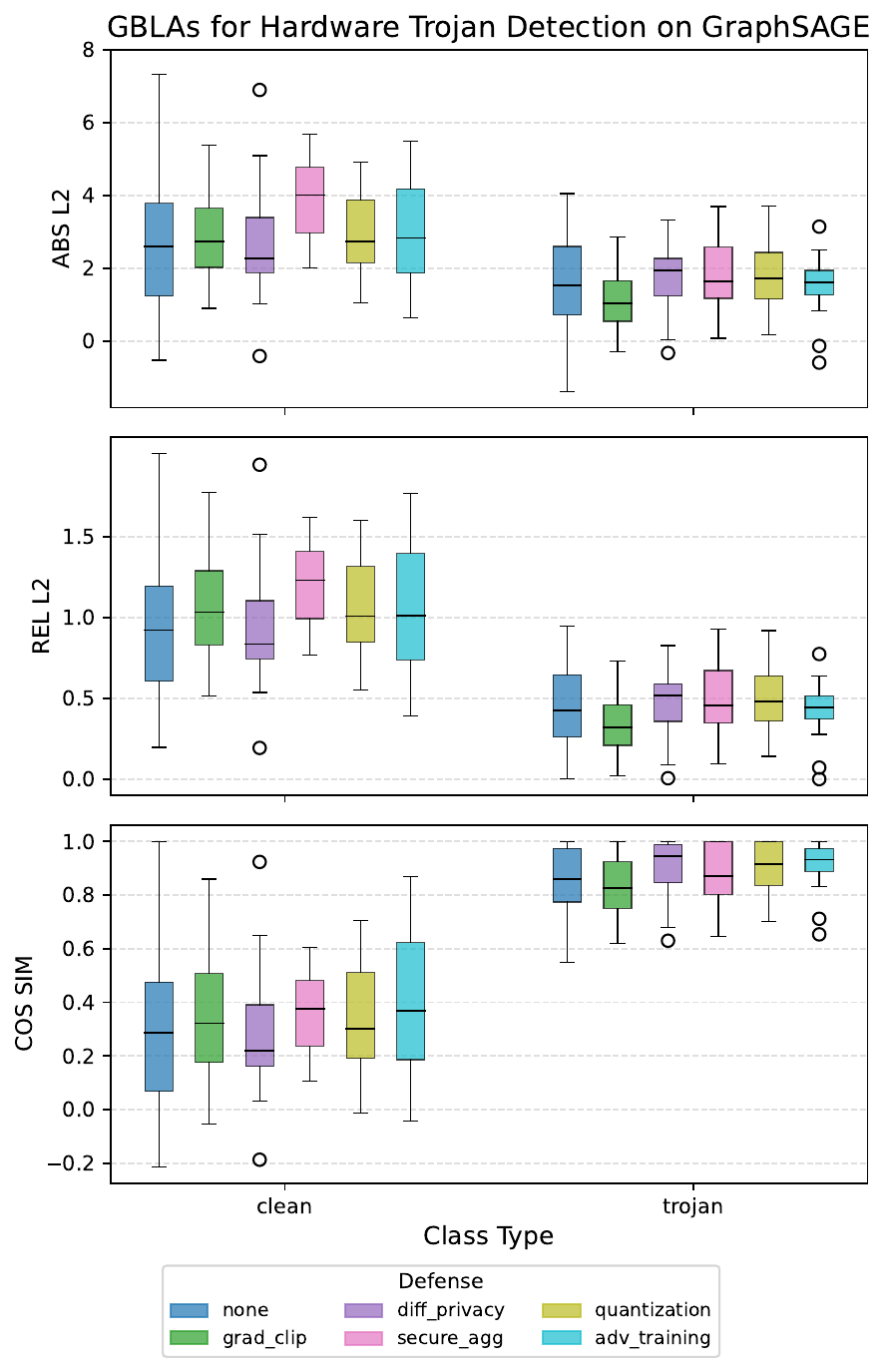} 
		\caption{\small Hardware Trojan Detection: Reconstruction errors under different defense mechanism for GraphSAGE. } 
		\label{fig:trojan_metrics_boxplots_GraphSAGE}
	\end{center}
\end{figure}

\begin{figure}[!t]
	\begin{center}
		\includegraphics[scale=0.58, trim = {0cm 0cm 0cm 0cm}, clip]{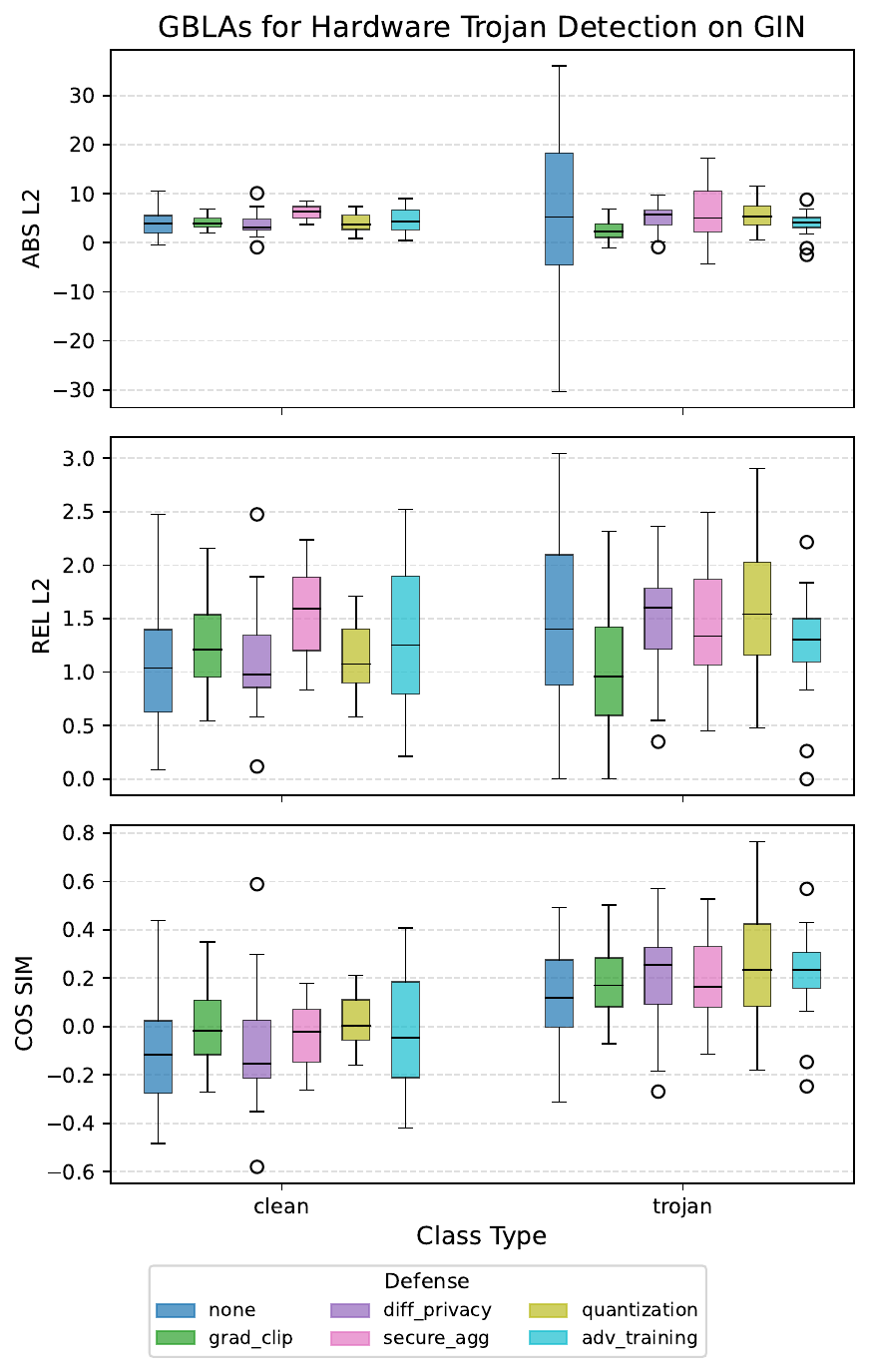} 
		\caption{\small Hardware Trojan Detection: Reconstruction errors under different defense mechanism for GIN. } 
		\label{fig:trojan_metrics_boxplots_GIN}
	\end{center}
\end{figure}

\begin{figure}[!t]
	\begin{center}
		\includegraphics[scale=0.58, trim = {0cm 0cm 0cm 0cm}, clip]{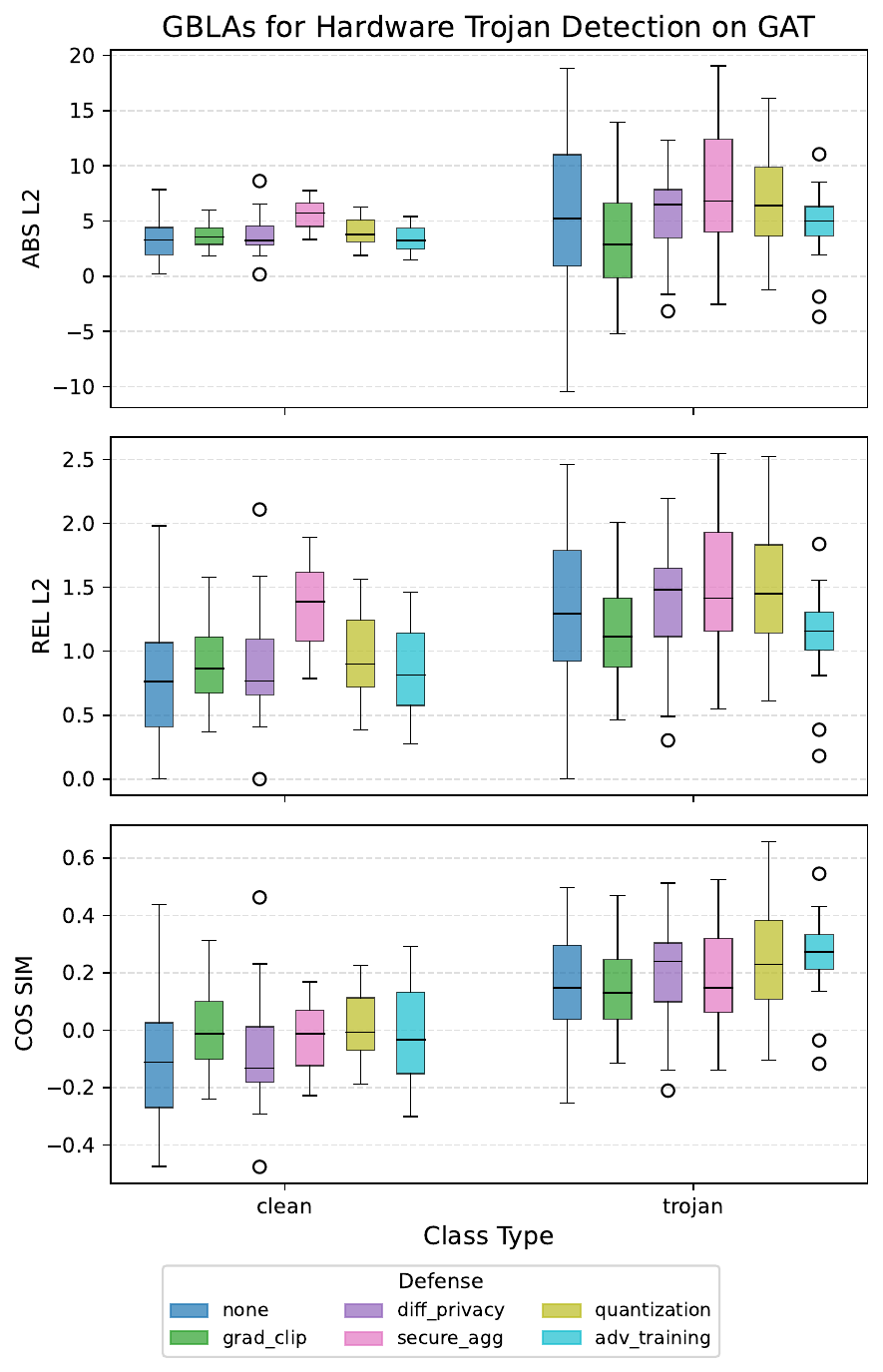} 
		\caption{\small Hardware Trojan Detection: Reconstruction errors under different defense mechanism for GAT. } 
		\label{fig:trojan_metrics_boxplots_GAT}
	\end{center}
\end{figure}

\subsection{Gradient Clipping and Perturbation}
\label{subsec:clip_perturbation_defense}

Here, we implement a defense mechanism that combines gradient clipping with perturbation. Let 
\[
G^{(l)}_{\text{true}} = \nabla_{W^{(l)}} \mathcal{L}(\hat{Y}, Y)
\]
denote the true gradient of the loss \(\mathcal{L}\) with respect to the model parameters \(W^{(l)}\) at layer \(l\). We first clip these gradients to restrict their magnitudes by enforcing
\[
G^{(l)}_{\text{clip}} = \min\left(1, \frac{C}{\| G^{(l)}_{\text{true}} \|_2} \right) \, G^{(l)}_{\text{true}},
\]
where the clipping threshold is set to \(C=1.0\). To further obscure the leaked information, we add Gaussian noise to produce the defended gradient:
\[
G^{(l)}_{\text{def}} = G^{(l)}_{\text{clip}} + \mathcal{N}(0, \alpha^2 I),
\]
with \(\alpha = 0.05\) and \(I\) being the identity matrix. The reconstruction objective then becomes
\[
\min_{\tilde{X}} \; \sum_{l=0}^{L-1} \left\| \nabla_{W^{(l)}} \mathcal{L}(\tilde{X}) - G^{(l)}_{\text{def}} \right\|_{F}^{2},
\]
where \(\tilde{X}\) is the dummy input being optimized. This two-fold defense -- clipping gradients to a maximum norm of 1.0 and perturbing them with a noise multiplier of 0.05 -- effectively degrades the precision of the gradient signal available to the attacker, leading to higher reconstruction errors and improved privacy protection.

\subsection{Secure Aggregation Protocols}
\label{subsec:secure_agg_defense}

Here, we apply secure aggregation protocols, wherein gradients are computed and averaged over multiple nodes instead of individual samples. This aggregated approach limits the attacker's ability to reconstruct specific node features while preserving the integrity of model updates. 

Let a group of \(N_{\text{group}}\) nodes be selected from a given class, each with its corresponding feature \(X_i\). The aggregated feature for the group is computed as:
\[
X_{\text{agg}} = \frac{1}{N_{\text{group}}} \sum_{i=1}^{N_{\text{group}}} X_i.
\]
Similarly, the aggregated gradients for the group are computed using the mean loss over all selected nodes:
\[
G_{\text{agg}}^{(l)} = \frac{1}{N_{\text{group}}} \sum_{i=1}^{N_{\text{group}}} \nabla_{W^{(l)}} \mathcal{L}(X_i, Y_i),
\]
where \(\mathcal{L}(X_i, Y_i)\) represents the loss function for node \(i\).

In our implementation, we set \(N_{\text{group}} = 5\) and simulate an attacker attempting to reconstruct \(X_{\text{agg}}\) rather than individual node features. The reconstruction objective now becomes:
\[
\min_{\tilde{X}_{\text{agg}}} \; \sum_{l=0}^{L-1} \left\| \nabla_{W^{(l)}} \mathcal{L}(\tilde{X}_{\text{agg}}) - G_{\text{agg}}^{(l)} \right\|_F^2.
\]

Since the attacker is only provided with the averaged gradient and feature information, individual variations within the group are masked, notably degrading inversion performance.

\subsection{Model Compression and Quantization}
\label{subsec:compression_quantization_defense}

Here, we use model compression and quantization techniques to restrict the precision of model parameters, obfuscating critical information that could be exploited by an attacker.
This approach involves two key defenses: (1) quantization, where model outputs are discretized to lower precision, and (2) compression through weight pruning, where small-valued parameters are removed to minimize information leakage.

\subsubsection{Quantization} 
During the forward pass of the GCNN, numerical outputs are quantized using:
\[
X_{\text{quant}} = \frac{\text{round}(X \cdot Q)}{Q},
\]
where \(Q = 255\) simulates 8-bit fixed-point representation. This transformation limits the granularity of feature updates, reducing inversion fidelity.

\subsubsection{Compression via Pruning}
After backpropagation, small-magnitude model weights are set to zero:
\[
W_{\text{compressed}}^{(l)} = 
\begin{cases} 
0, & \text{if } |W^{(l)}| < \tau \\ 
W^{(l)}, & \text{otherwise}
\end{cases}
\]
where \(\tau = 0.01\) is the pruning threshold. This process eliminates weak connections in the NN, making gradient inversion less effective.

\subsection{Adversarial Training}
\label{subsec:adv_training_defense}

Here, we introduce adversarial perturbations into the model’s training process to improve its resilience against GBLAs. By exposing the model to carefully crafted perturbations during training, it learns representations that are less susceptible to inversion attacks. 

\subsubsection{Generating Adversarial Perturbations}
To generate adversarial examples, we use the Fast Gradient Sign Method (FGSM), where the perturbation applied to the feature matrix \(X\) is computed as:
\[
X_{\text{adv}} = X + \epsilon \cdot \text{sign}(\nabla_X \mathcal{L}(X, Y)),
\]
where \(\epsilon\) is the perturbation strength (set to \(0.1\) in our experiments), \(\mathcal{L}(X, Y)\) is the loss function, and \(\nabla_X \mathcal{L}\) represents the gradient of the loss with respect to the input features.

\subsubsection{Adversarial Training Procedure}
During training, instead of using the original features, we compute adversarial perturbations and feed the model with \(X_{\text{adv}}\), reinforcing its ability to generalize across perturbed data:
\[
\min_W \sum_{i=1}^{N} \mathcal{L}(f_W(X_{\text{adv}, i}), Y_i).
\]
This forces the model to optimize for robustness rather than merely fitting the clean training data.

\section{Plots for Different Defenses under GNNs}

The box plots for different metrics \texttt{(abs\_l2, rel\_l2, and cos\_sim)} for both the gate classification and the hardware Trojan detection on GCN, GraphSAGE, GIN, and GAT models are shown in Figs.~\ref{fig:gate_metrics_boxplots_combined_gcn}, \ref{fig:gate_metrics_boxplots_combined_graphsage}, \ref{fig:gate_metrics_boxplots_combined_GIN}, \ref{fig:gate_metrics_boxplots_combined_GAT}, \ref{fig:trojan_metrics_boxplots_GraphSAGE}, \ref{fig:trojan_metrics_boxplots_GIN}, and \ref{fig:trojan_metrics_boxplots_GAT} respectively.

\end{document}